\documentclass[lettersize,journal]{IEEEtran}
\usepackage[backend=biber, style=numeric-comp]{biblatex}
\usepackage{amsmath,amsfonts}
\usepackage{algorithmic}
\usepackage{algorithm}
\usepackage{array}
\usepackage{textcomp}
\usepackage{stfloats}
\usepackage{url}
\usepackage{verbatim}
\usepackage{booktabs}
\usepackage{multirow}
\usepackage[table]{xcolor}    % 关键宏包：支持表格颜色（[table]选项必须）
\usepackage{array}            % 增强表格格式控制
\usepackage{lipsum}
\usepackage{graphicx}
\usepackage{float}  %设置图片浮动位置的宏包
\usepackage{caption}
\definecolor{MyColor1}{RGB}{255 ,0, 0}
\definecolor{MyColor2}{RGB}{0, 255, 0}
\definecolor{MyColor3}{RGB}{0, 0, 255}
\definecolor{MyColor4}{RGB}{255, 255, 0}
\definecolor{MyColor5}{RGB}{255, 0, 255}
\definecolor{MyColor6}{RGB}{0, 255, 255}
\definecolor{MyColor7}{RGB}{200, 100, 0}
\definecolor{MyColor8}{RGB}{0, 200, 100}
\definecolor{MyColor9}{RGB}{100, 0, 200}
\definecolor{MyColor10}{RGB}{200, 0, 100}
\definecolor{MyColor11}{RGB}{100, 200, 0}
\definecolor{MyColor12}{RGB}{0, 100, 200}
\definecolor{MyColor13}{RGB}{150, 75, 75}
\definecolor{MyColor14}{RGB}{75, 150, 75}
\definecolor{MyColor15}{RGB}{75, 75, 150}
\definecolor{MyColor16}{RGB}{255, 100, 100}
\definecolor{MyColor17}{RGB}{100, 255, 100}
\definecolor{MyColor18}{RGB}{100, 100, 255}
\definecolor{MyColor19}{RGB}{255, 150, 75}
\definecolor{MyColor20}{RGB}{75, 255, 150}
\definecolor{MyColor21}{RGB}{150, 75, 255}
\definecolor{MyColor22}{RGB}{50, 50, 50}

\usepackage{subcaption}
\usepackage[
colorlinks,
linkcolor=black,
anchorcolor=black,
citecolor=black
]{hyperref}

\usepackage{orcidlink}
\usepackage{placeins}
\begin{document}

\title{HS-Mamba: Full-Field Interaction Multi-Groups Mamba for Hyperspectral Image Classification}

\author{
\thanks{This research was supported by Natural Science Foundation of Guangdong Province (Grant No. 2025A1515011771) and Guangzhou Science and Technology Plan Project (Grant No. 2023B01J0046, 2024E04J1242).}
Hongxing Peng$^{\orcidlink{0000-0002-1872-8855}}$\
\thanks{Hongxing Peng are with the College of Mathematics and Informatics, South China Agricultural University, Key Laboratory of Smart Agricultural Technology in Tropical South China, Ministry of Agriculture and Rural Affairs, Guangzhou 510642, China (e-mail:xyphx@scau.edu.cn).},
Kang Lin$^{\orcidlink{0009-0005-7175-4752}}$\thanks{Kang Lin are with the College of Mathematics and Informatics, South China Agricultural University, Guangzhou 510642, China (e-mail:lkang4702@gmail.com).},
and Huanai Liu$^{\orcidlink{0009-0001-8257-0570}}$\thanks{Huanai Liu are with the School of Chemistry and Chemical Engineering, South China University of Technology, Guangzhou 510641, China (e-mail: liuhn@scut.edu.cn).}
}


\maketitle
\begin{abstract}
Hyperspectral image (HSI) classification has been one of the hot topics in remote sensing fields. Recently, the Mamba architecture based on selective state-space models (S6) has demonstrated great advantages in long sequence modeling. However, the unique properties of hyperspectral data, such as high dimensionality and feature inlining, pose challenges to the application of Mamba to HSI classification. To compensate for these shortcomings, we propose an full-field interaction multi-groups Mamba framework (HS-Mamba), which adopts a strategy different from pixel-patch based or whole-image based, but combines the advantages of both. The patches cut from the whole image are sent to multi-groups Mamba, combined with positional information to perceive local inline features in the spatial and spectral domains, and the whole image is sent to a lightweight attention module to enhance the global feature representation ability. Specifically, HS-Mamba consists of a dual-channel spatial-spectral encoder (DCSS-encoder) module and a lightweight global inline attention (LGI-Att) branch. The DCSS-encoder module uses multiple groups of Mamba to decouple and model the local features of dual-channel sequences with non-overlapping patches. The LGI-Att branch uses a lightweight compressed and extended attention module to perceive the global features of the spatial and spectral domains of the unsegmented whole image. By fusing local and global features, high-precision classification of hyperspectral images is achieved. Extensive experiments demonstrate the superiority of the proposed HS-Mamba, outperforming state-of-the-art methods on four benchmark HSI datasets.
\end{abstract}


\section{Introduction}
\IEEEPARstart{H}{yperspectral} image (HSI) is an advanced remote sensing technique that captures the electromagnetic wave reflection characteristics of land objects through continuous and dense narrow bands. It can form continuous spectral curves in dozens or even hundreds of bands ranging from visible light to near-infrared, accurately characterizing the unique spectral characteristics of different substances\cite{HSIDataAnalysis}. Due to the rich spectral bands and the interconnectivity between spatial and spectral domains, HSI classification has become a fundamental research in the field of remote sensing, playing an important role in tasks such as urban surveying\cite{urban}, environmental exploration\cite{environment}, and precision agriculture\cite{agriculture}.

\begin{figure}
\centering
\includegraphics[width=3.6in, keepaspectratio]{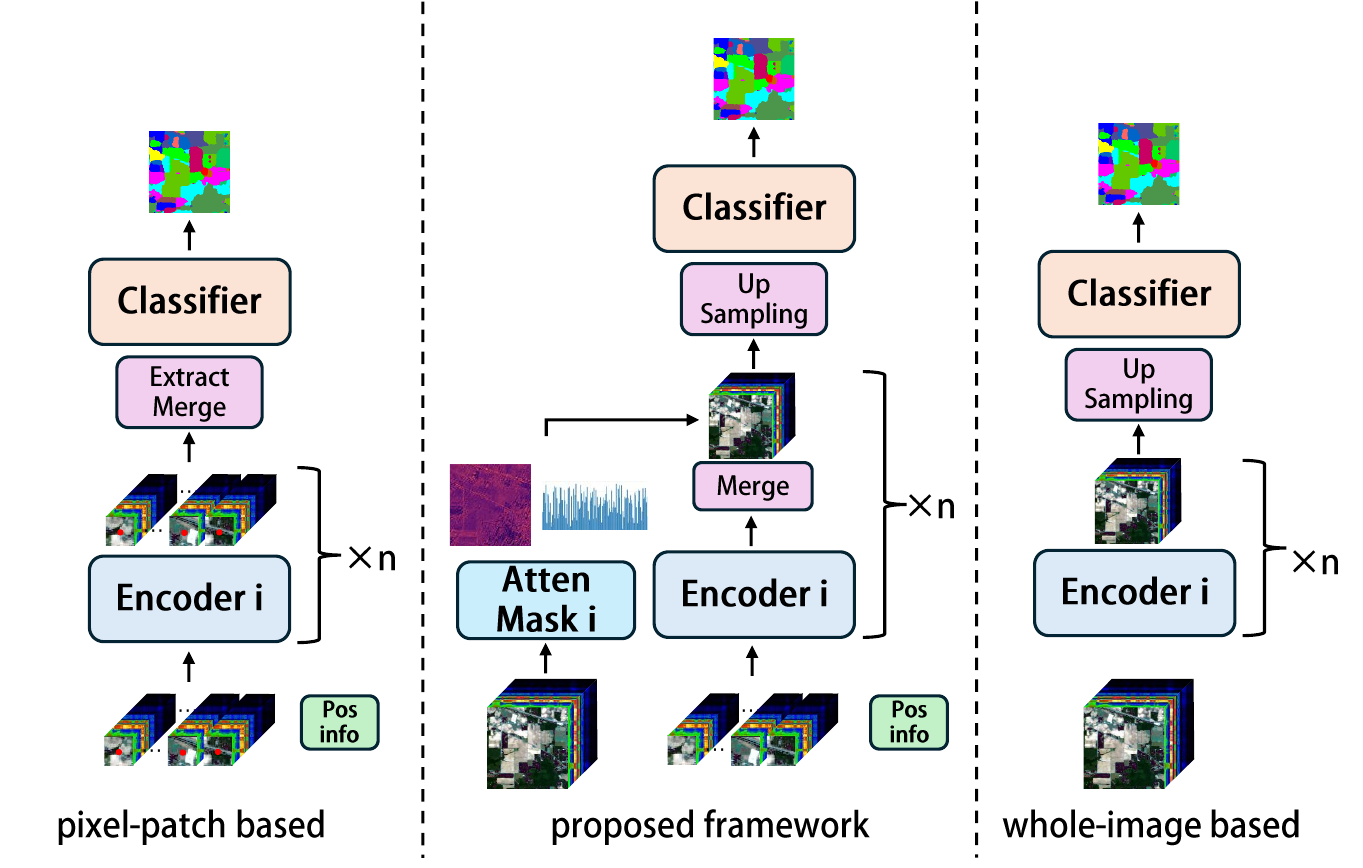}\\
\caption{Innovation of strategy. The traditional pixel-patch based strategy and the whole-image based strategy have their own defects: the pixel-patch based strategy lacks the understanding of the global semantics and and suffers from pixel noise sensitivity, while the whole-image based strategy has the problem of losing high-frequency local details. The proposed full-field interaction strategy strikes an optimal balance between the two and achieve efficient full-field representation.}
\label{fig:innovation of strategy}
\end{figure}

The ultimate goal of HSI classification is to accurately assign each pixel to the corresponding surface classification\cite{purpose}. Classification methods are specifically divided into Machine Learning (ML) and Deep Learning (DL). Among numerous ML-based methods, researchers have explored the characteristics of multidimensional spectra and adopted ML-based methods\cite{SVM}\cite{RF}, combined with dimensionality reduction methods\cite{PCA}\cite{LDA} to better address the spectral redundancy and pixel noise characteristics of HSI datasets\cite{noice}. Some researchers have focused on exploring the inline features between spatial and spectral domains, and developed manually designed frameworks\cite{EMAP}\cite{SMP} to achieve classification. Obviously, these ML-based methods heavily rely on prior knowledge, and the extremely high design threshold and lack of universality in complex environments make them difficult to generalize\cite{hardToDesign}.

The DL-based methods can be divided into CNN-based\cite{CNNHSI} and Transformer-based\cite{TransformerHSI} methods. The translation invariance of CNN\cite{1D-CNN} and the powerful modeling ability of Transformer for long sequences\cite{goodTransformer} make them widely accepted. However, CNN-based methods are limited by their local receptive fields and cannot effectively establish a global perspective\cite{badCNN}. Although Transformer-based methods have excellent modeling capability for long sequences, the quadratic complexity for sequence length makes them not computationally advantageous and cannot be widely used in environments with limited computing resources\cite{Transformer}. Therefore, it is urgent to propose a model that can strike a balance between long-distance modeling capability and computational complexity.

Recently, the Structured State Space Model (S4)\cite{S4} has demonstrated excellent capability in modeling continuous long sequence data and has been applied in numerous deep network constructions. Based on S4, Mamba\cite{Mamba} integrates a selective scanning mechanism, allowing the model to dynamically adjust parameters based on inputs. At the same time, the addition of hardware aware algorithms significantly improves the computational efficiency of the model, with a much lower computational complexity than Transformers\cite{Mamba}. This outstanding feature of Mamba in long-range language modeling while maintaining linear computational complexity has led to its application in many Natural Language Processing tasks. At the same time, many studies have begun to explore the use of Mamba models for Computer Vision tasks\cite{Vim}\cite{VMamba}, which deliver competitive results in classification, detection, and segmentation. However, due to the spatial spectral inlining features of HSI data and the complexity of the spectrum, Mamba is hard to directly apply to HSI classification tasks\cite{MambaHSI}.

In the past two years, Mamba-based HSI classification models have emerged in an endless stream. According to the strategy, they can be divided into two categories, namely pixel-patch based classification\cite{DualMamba} and whole-image based classification\cite{MambaHSI}, each with its own advantages. However, both methods have insurmountable challenges, as shown in Fig. \ref{fig:innovation of strategy}.
\begin{itemize}
    \item [1)]The pixel-patch based model tends to ignore global semantic information and is greatly affected by pixel noise. Expanding a single pixel to a patch input model means that there is high-frequency repeated calculation at each position, wasting computing resources. 
\end{itemize}
\begin{itemize}
    \item [2)]The whole-image based model is difficult to overcome the problem of losing high-frequency local details, and it is easy to cause GPU-memory overflow on high-resolution HSI images.
\end{itemize}
 
In order to meet the above challenges, we propose a full-field interaction multi-groups Mamba framework (HS-Mamba) for HSI classification. This framework combines the advantages of pixel-patch based and whole-image based classification. We process non-overlapping image patches with positional encodings through parallel Mamba groups, effectively avoiding redundant computations while harnessing their capability to extract local spatial-spectral patterns. Simultaneously, the full-resolution image is fed into a lightweight attention module that incorporates comprehensive global context to strengthen feature learning across the entire scene. Specifically, HS-Mamba consists of a dual-channel spatial-spectral encoder (DCSS-encoder) module and a lightweight global inline attention (LGI-Att) branch. The DCSS-encoder module uses multi-groups Mamba and adaptive concat, to decouple and model the local information of the 1D flattened spatial and spectral domain sequences of non-overlapping patches. The LGI-Att branch uses a lightweight compression and expansion attention module to perceive the global features of the spatial and spectral domains of the whole-image. The two branches extract local and global features respectively, and then fuse the attention features to achieve high-precision classification of HSI. Our main contributions are summarized as follows.
\begin{itemize}
    \item [1)] In terms of strategy, we propose HS-Mamba. This is a classification strategy different from pixel patch-based and whole image-based. The framework combines the non-overlapping patches and the whole-image to achieve the attention of local inline relationships while enhancing the model's global feature representation capability.
\end{itemize}
\begin{itemize}
    \item [2)] We propose the DCSS-Encoder module, which integrates multi-group Mamba with an adaptive concatenation strategy and cosine positional encoding to model local features in 1D flattened spatial-spectral sequences derived from non-overlapping patches. This design enables comprehensive analysis of spectral-spatial relationships while maintaining a global perspective.
\end{itemize}
\begin{itemize}
    \item [3)] We proposed the LGI-Atten branch, which uses lightweight compressed attention for global information in the spatial domain and lightweight extended attention for global feature perception of spectral information. These attention information will be converted into global weight information and combined with the trunk branch.
\end{itemize}
\begin{itemize}
    \item [4)] Comprehensive experiments conducted on four widely recognized HSI classification benchmark datasets, namely Indian Pines, Pavia University, WHU-Hi-HanChuan, and WHU-Hi-HongHu, show that our method achieves state-of-the-art (SOTA) performance.
\end{itemize}

\section{Related work}
\subsection{HSI Classification}
HSI classification methods are generally divided into ML-based\cite{SVM}\cite{EMAP} and DL-based methods\cite{CNNHSI}\cite{TransformerHSI}. ML-based methods are roughly divided into traditional ML-based methods\cite{SVM}\cite{RF} and methods manually designed based on prior knowledge\cite{EMAP}\cite{SMP}. DL-based methods are roughly divided into CNN-based\cite{CNNHSI}\cite{badCNN} and Transformer-based methods\cite{TransformerHSI}. We discuss CNN-based, Transformer-based, and traditional ML-based methods in parallel due to their fundamental architectural differences.

\subsubsection{ML-based}
Early ML-based methods for HSI analysis focused on exploiting intrinsic data properties through shallow models like Support Vector Machine (SVM)\cite{SVM} and Random Forest (RF)\cite{RF}, complemented by dimensionality reduction techniques such as Principal Component Analysis (PCA)\cite{PCA} and Linear Discriminant Analysis (LDA)\cite{LDA} to address high dimensionality and noise\cite{noice}. However, these shallow models exhibit limited representation capacity, while dimensionality reduction risks critical information loss\cite{MambaHSI}. Subsequent work explored spatial-spectral joint representations through manually designed frameworks like Extended Multi Attribute Graph (EMAP)\cite{EMAP} and sparse manifold methods\cite{SMP}. Despite their theoretical value, these approaches require substantial domain expertise for feature engineering and demonstrate poor generalization across complex scenarios\cite{hardToDesign}, significantly limiting practical applicability.

\subsubsection{CNN-based}
Advancing hardware and AI cognition have propelled DL-based methods, particularly CNN architectures with adaptive receptive fields and scale invariance, into prominence for HSI classification through continuous innovation. The researchers first focused on the shape of HSI and successively proposed 1D-CNN\cite{1D-CNN}, 2D-CNN\cite{2D-CNN} and 3D-CNN\cite{3D-CNN}, examining the spatial domain and high-dimensional frequency domain information from 1D to 3D perspective. As the information increased, the classification effect continued to improve. But obviously, 3D-CNN is difficult to overcome the problem of large number of parameters. Hence, Li et al.\cite{CLOLN-CNN} proposed a channel-layer-oriented lightweight network, which uses single and double channel 3D-CNN and combines deep-width convolution to achieve the purpose of lightweight. However, the local receptive field of convolution makes it inevitable to have deficiencies in processing context information\cite{MambaHSI}. Therefore, Xu et al.\cite{FirstImageHSI-CNN} first proposed to input the entire image directly into the model instead of inputting pixel patches one by one to learn global representations. Wang et al.\cite{FullyContext-CNN} followed closely and proposed a FullyContNets, which is also based on the entire image and adaptively aggregates multiple features through a pyramid multi-scale structure. Experiments have shown that good HSI classification results can be achieved through rich location information from the entire image.

\subsubsection{Transformer-based}
The rise of Transformer architectures\cite{Transformer}, enabled by their superior long-range modeling capabilities compared to CNNs, has revolutionized HSI classification research\cite{goodTransformer}. Hong et al.\cite{SF-Transformer} proposed a SpectralFormer based on pure Transformer to learn long-range sequence features from adjacent bands. Liao et al. \cite{TransformerHSI} also used multi-head attention to completely replace CNN, greatly reducing the number of parameters while achieving good results. Wang et al.\cite{SSFTT-Transformer} proposed a SSFTT that combines CNN and Transformer to capture spectral-spatial features and high-level semantic features from shallow to deep, while taking advantage of the advantages of CNN and Transformer. Peng et al.\cite{CASST-Transformer} proposed a spatial-spectral Transformer with cross-attention learning to model semantic features by fully learning contextual information. Chen et al.\cite{SQSformer-Transformer} re-examined the role of center pixels in HSI classification and proposed an SQSFormer integrated with a rotation-invariant embedding module, focusing on the center pixel, reducing the introduction of unnecessary information, and improving classification performance. Zhao et al.\cite{GSC-ViT} noticed the inherent high parameter calculation of Transformer and the neglect of local feature representation, and thus proposed a lightweight ViT network GSC-ViT, combining grouped separable convolutions and skip connections, which also improved classification performance. Although researchers have begun to improve Transformer-based methods in the direction of lightweightness, its inherent quadratic complexity hinders the upper limit of the model's application\cite{VMamba}.

\subsection{State Space Models and Mamba}
The state space model (SSM)\cite{SSM1}\cite{SSM2} originated from Kalman filter can construct a decoupled representation of a dynamic system through first-order derivatives and vectors.  On this basis, the structured SSM (S4)\cite{S4} introduces a structured state transfer matrix, which has advantages in long-range modeling. Mamba\cite{Mamba} introduces the selective structured SSM (S6) and integrates hardware-aware algorithms, which can achieve more efficient calculations while achieving long-range representation based on input dynamics\cite{VMamba}.

\subsubsection{Mamba for computer vision}
The success of Mamba in long-sequence language modeling has spurred its adaptation to computer vision, where most implementations convert 2D images into 1D sequences. Zhu et al.\cite{Vim} first proposed a Vision Mamba (ViM) architecture, which uses bidirectional scanning to process image data, allowing Mamba to perceive global features. Liu et al.\cite{VMamba} subsequently proposed a VMamba framework based on cross-select scanning, which uses four-way scanning to fully perceive spatial features. Based on the above research, Mamba has made continuous research in visual fields such as medicine\cite{medicine-Mamba}, agriculture\cite{agriculture-Mamba} and remote sensing\cite{remoteSensing-Mamba}.

\subsubsection{Mamba for HSI classification}
Recently, Mamba-based HSI classification methods are being explored. Huang et al.\cite{SS-Mamba} proposed a SS-Mamba based on spatial-spectral stacked Mamba blocks, and sent the generated tokens into the stacked Mamba structure for classification. In view of the 3D strong correlation characteristics of HSI data, He et al.\cite{3DSS-Mamba} proposed a spectral-spatial Mamba based on 3D scanning, and achieved good results. Zhou et al.\cite{Mamba-in-Mamba} proposed a classification framework with a centralized scanning strategy, which can transform sequences with a new scanning strategy. All of the above are innovations based on scanning strategies. This multi-directional scanning strategy poses a challenge to computational efficiency\cite{DualMamba}. He et al.\cite{IGroupSS-Mamba} proposed a lightweight multi-scale scanning Mamba based on interval grouping for classification. However, interval grouping may destroy the inline relationship of adjacent spectra. Li et al.\cite{MambaHSI} focused on the whole-image features and proposed the first pure Mamba framework based on the whole-image input, which captures dual-domain features through spectral and spatial dual branches, and then adaptively fused them to achieve ideal classification results. However, strategies based on the whole-image tend to ignore high-frequency local features, and the whole-image input poses a challenge to GPU memory.

To this end, we propose a full-field interaction multi-groups Mamba framework (HS-Mamba), which combines the non-overlapping patch and the whole-image to achieve the attention of local inline relationships while enhancing the model's global feature representation capability.

\section{Preliminaries}
The state space model (SSM)\cite{SSM1}\cite{SSM2} originated from Kalman filtering and its various variants (S4\cite{S4}, Mamba\cite{Mamba}) are all designed to establish a linear mapping relationship from the input one-dimensional signal $x(t) \in \mathbb{R}$ to the output signal $y(t) \in \mathbb{R}$ through the intermediate hidden state $h(t) \in \mathbb{R}^{N}$. This process can be represented by the following linear ordinary differential equation (ODE)
\begin{equation}
\begin{aligned}
h^{\prime}\left(t\right) & =\mathbf{A}h\left(t\right)+\mathbf{B}x\left(t\right), \\
y\left(t\right) & =\mathbf{C}h\left(t\right)
\end{aligned}
\end{equation}
where $A \in \mathbb{R}^{N\times N}$ denotes the state matrix, and  $B \in \mathbb{R}^{N\times 1}$  and $C \in \mathbb{R}^{N\times 1}$ signify the projection parameters.

In order to enable SSM to be integrated into the discrete-sequences model, researchers use the zero-order hold (ZOH) technique with a time scale parameter $\Delta$ to transform continuous parameters $\textit{A}$ and $\textit{B}$ into discrete parameters $\overline{A}$ and $\overline{B}$. This process is defined as follows:
\begin{equation}
\begin{aligned}
 & \overline{A}=\exp(\Delta\mathbf{A}), \\
 & \overline{B}=(\Delta\mathbf{A})^{-1}(\exp(\Delta\mathbf{A})-\mathbf{I})\cdot\Delta\mathbf{B}
\end{aligned}
\end{equation}

After discretizing the continuous parameters with input size $\textit{L}$, the researchers optimized the training and evaluating process based on the ideas of recurrent and convolutional operator $\overline{\mathbf{K}} \in \mathbb{R}^{L}$, which represents a convolution kernel with size $\textit{L}$. The overall calculation process can be defined as follows:
\begin{equation}
\begin{aligned}
 & h_{t}=\overline{\mathbf{A}}h_{t-1}+\overline{\mathbf{B}}x_t, \\
 & y_{t}=\mathbf{C}h_t, \\
 & \overline{\mathbf{K}}=\left(\mathbf{C\overline{B}},\mathbf{C\overline{AB}},\ldots,\mathbf{C\overline{A}}^{L-1}\mathbf{\overline{B}}\right), \\
 & =x*\overline{\mathbf{K}}
\end{aligned}
\end{equation}

All SSM-based methods are inseparable from the above architecture. Subsequently, S4 added structured representation based on discretization, and Mamba optimized the selective dynamic input and hardware calculation mechanism based on S4, which made the model go a step further.

\section{Methodology}
\begin{figure*}
% \subfigbottomskip=2pt %两行子图之间的行间距
% \subfigcapskip=-1pt %设置子图与子标题之间的距离
\centering
\begin{subfigure}{.9\textwidth}
\centering
% \subfigure[Overview Framework of the proposed HS-Mamba]{
\includegraphics[width=1\linewidth]{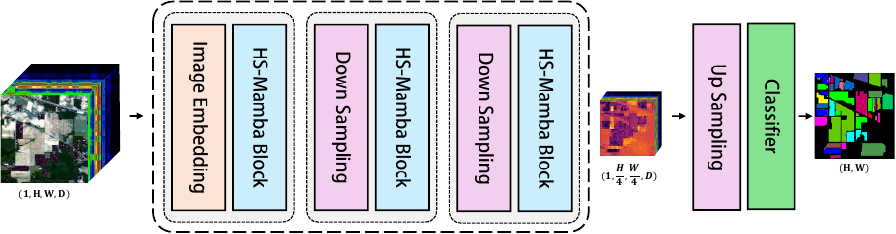}
\caption{Overview Framework of the proposed HS-Mamba}
\label{fig:overview architecture}
% }
\end{subfigure}
\newline
\begin{subfigure}{.4\textwidth}
\centering
% \subfigure[Specific Scanning Process of Spectral and Spatial Scan]{
\includegraphics[width=0.92\linewidth]{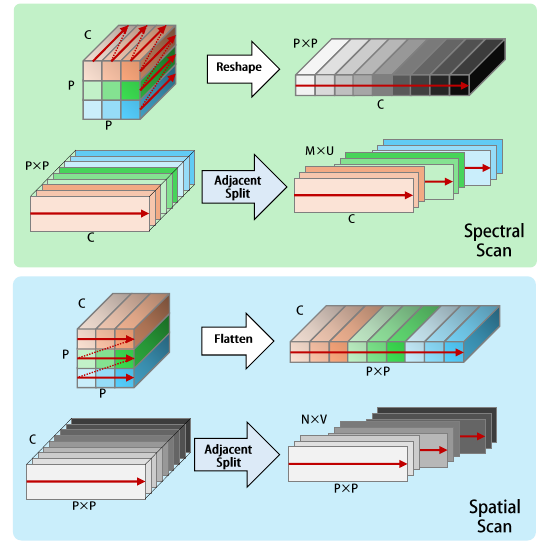}
\caption{Specific Scanning Process of Dual-domain Scan}
\label{fig:scan}
% }
\end{subfigure}
\begin{subfigure}{.5\textwidth}
\centering
% \subfigure[The Detailed Structure of HS-Mamba Block]{
\includegraphics[width=0.96\linewidth]{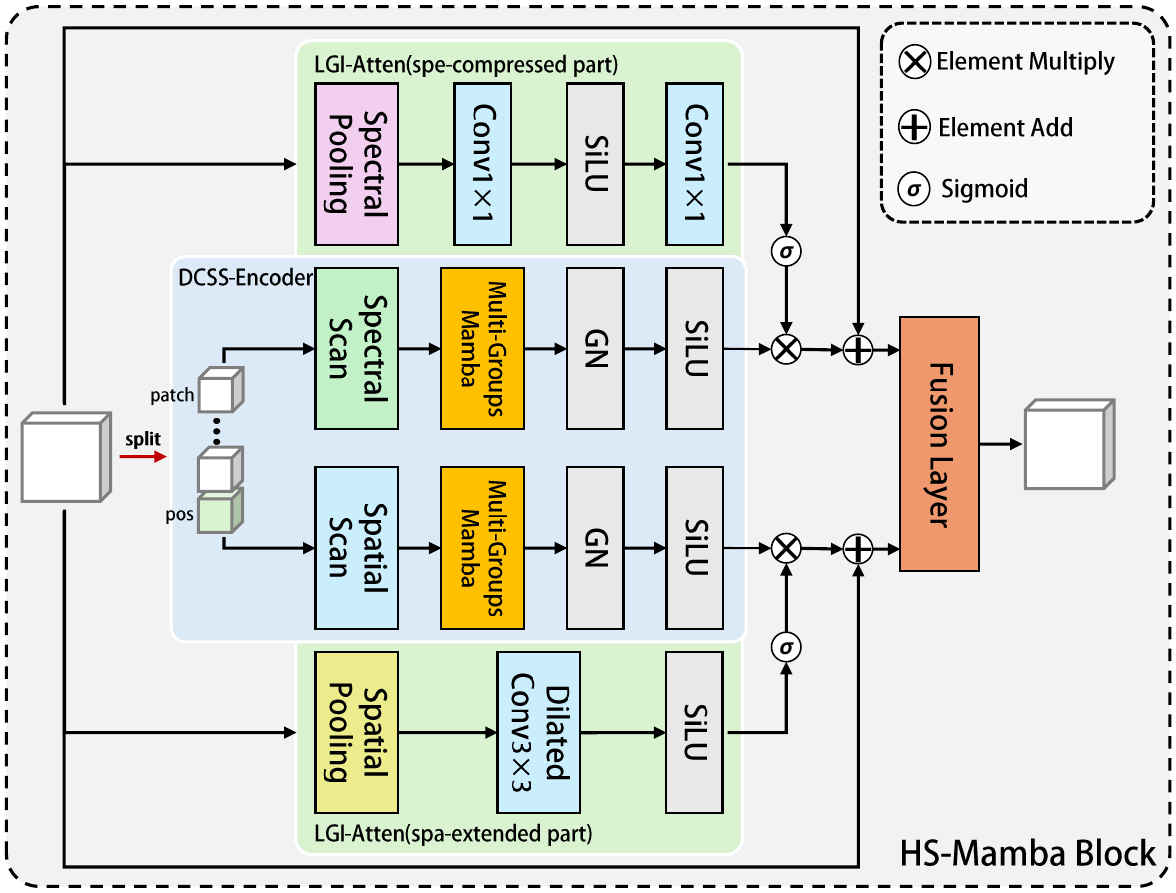}
\caption{The Detailed Structure of HS-Mamba Block}
\label{fig:block}
% }
\end{subfigure}
\caption{The overview of the proposed full-field interaction multi-groups Mamba (HS-Mamba) for HSI classification. (a) The overall architecture of the proposed HS-Mamba, three-stage architecture employs HS-Mamba blocks for hierarchical representation, achieving refined classification through two down-sampling layers and a up-sampling operation; (b) The computational procedure of the proposed dual-domain scanning for spatial and spectral features; (c) The HS-Mamba Block processes HSI input through parallel pathways: the LGI Attention module extracts global features while the DCSS-Encoder processes local position-enhanced non-overlapping patches, followed by gated fusion to yield the integrated representation.}
\label{fig:overview}
\end{figure*}

\subsection{Overview Framework}
In this paper, we propose HS-Mamba, a novel hyperspectral image classification framework that integrates pixel-patch and whole-image analysis strategies to achieve state-of-the-art (SOTA) accuracy. As shown in Fig. \ref{fig:overview architecture}, given an input HSI cube $I\in \mathbb{R}^{1\times C\times H\times W}$ with corresponding labels $l\in \mathbb{R}^{1\times H\times W}$. $\textit{H}$ and $\textit{W}$ represent the height and width of the HSI image data, and $\textit{C}$ represents the band length. The framework operates through three hierarchical stages. Each stage contains an HS-Mamba Block that progressively extracts joint spectral-spatial features from shallow to deep representations, ultimately enabling precise pixel-level classification via up-sampling and a dedicated classifier.

\subsubsection{Image Embedding}
The process is used to preliminarily extract shallow HSI features and unify the subsequent processing dimension. Specifically, this process can transform the original HSI cube data $I\in \mathbb{R}^{1\times C\times H\times W}$ into a refined whole-image embedding ${F_{embed}}\in \mathbb{R}^{1\times D\times H\times W}$, mathematically expressed as
\begin{equation}
F_{\mathrm{embed}} =\operatorname{SiLU}(\operatorname{GN}(\operatorname{Conv_{1\times 1}}(I)))
\end{equation}
Where $\mathbf{D}$ represents the unified embedding channel dimension. $\operatorname{Conv_{1\times 1}}$ is used for preliminary processing.

% \subsubsection{Down and Up Sampling}
% To capture multi-scale feature representations and enhance inter-patch correlations, we employ average pooling for down-sampling operations. The original spatial resolution is subsequently restored through bilinear interpolation-based up-sampling, enabling precise pixel-level classification through the final classifier, mathematically expressed as
% \begin{equation}
% \begin{aligned}
% F_{\mathrm{down}}^i &= \operatorname{DownSampling}(F_{\mathrm{block}}^i),  && i = 1, 2, \\
% F_{\mathrm{up}}     &= \operatorname{UpSampling}(F_{\mathrm{encoder}})
% \end{aligned}
% \end{equation}
% where ${F_{\mathrm{block}}^i} \in \mathbb{R}^{1\times D\times {H^i}\times {W}^i}$ denotes the HS-Mamba block output, ${F_{encoder}} \in \mathbb{R}^{1\times D\times \frac{H}{2}\times\frac{W}{2}}$ denotes the encoder output, ${F_{\mathrm{down}}^i} \in \mathbb{R}^{1\times D\times \frac{{H}^i}{2}\times \frac{{W}^i}{2}}$ and ${F_\mathrm{up}} \in \mathbb{R}^{1\times D\times H\times W}$ denotes the down-sample and up-sampling layer output.

\subsubsection{Fusion Layer}
After obtaining dual-domain features, we propose a Gated Fusion strategy. Specifically, channel-wise concatenated features pass through a dimensionality reduction layer to derive weighting parameters, enabling adaptive weighted combination of both branches for comprehensive feature integration, mathematically expressed as
\begin{equation}
\begin{aligned}
w_\mathrm{gated} &= \operatorname{Conv_{1\times 1}}(\operatorname{Concat}(F_\mathrm{spe}, F_\mathrm{spa})), \\
F_\mathrm{fusion} &= w_\mathrm{gated}\times F_\mathrm{spe} + (1 - w_\mathrm{gated})\times F_\mathrm{spa}
\end{aligned}
\end{equation}
where $\operatorname{Conv_{1\times 1}}$ is used for dimension adjustment $\mathbb{R}^{2\times D} \to \mathbb{R}^{1}$, $F_{fusion} \in \mathbb{R}^{1\times D\times H\times W}$ denotes the output of the fusion layer.

\subsubsection{Classifier}
The classification head mathematically expressed as
\begin{equation}
Y =\operatorname{Conv^{2}_{1\times 1}}(\operatorname{SiLU}(\operatorname{GN}(\operatorname{Conv^{1}_{1\times 1}}(F_\mathrm{up}))))
\end{equation}
where ${Y} \in \mathbb{R}^{1\times \mathbf{num\_classes}\times H\times W}$ denotes the final classification output, $\operatorname{Conv^{1}_{1\times 1}}$ is used for feature learning and $\operatorname{Conv^{2}_{1\times 1}}$ is used for dimension adjustment $\mathbb{R}^{D} \to \mathbb{R}^\mathbf{num\_classes}$.

\subsection{Dual-Channel Spatial-Spectral Encoder}
Conventional pixel-patch based encoder suffer from inefficient inference due to redundant computations of overlapping neighborhood features, where adjacent patches repeatedly process shared spectral-spatial regions. To resolve this, we propose the DCSS-Encoder that strategically extracts non-overlapping patches while integrating positional encoding and multi-group Mamba modules to capture intra-patch dependencies without feature recomputation. The specific architecture is shown in Fig. \ref{fig:block}.

\subsubsection{Spectral and Spatial Scan}
Sequence modeling of 2D HSI data requires efficient 1D scanning strategies. Conventional multi-directional scanning approaches for HSI data obtained by remote sensing often introduce redundant spatial features and computational inefficiencies\cite{DualMamba}. Therefore, the unidirectional scanning strategy becomes the best choice. The specific architecture is shown in Fig. \ref{fig:scan}.

Specifically, given the input of non-overlapping HSI patch tokens $F_\mathrm{patch} = \{{S_1}, {S_2},...,{S_N}\}$, ${S_i} \in \mathbb{R}^{1\times D\times P\times P}$, where P and L denotes the patch size and the patch count. $\mathrm{P\times P\times L}$ is equal to $\mathrm{H\times W}$. The spectral domain employs spectral-priority scanning to organize spectral units followed by pixel-wise arrangement, while spatial-domain scanning prioritizes pixel units before spectral sequencing, mathematically expressed as
\begin{equation}
\begin{aligned}
{S_i}^\mathrm{spe} &= [[S_{{i,1}}^{1},...,S_{{i,1}}^{D}],...,[S_{{i,{P^2}}}^{1},...,S_{{i,{P^2}}}^{D}]], \\ 
{S_i}^\mathrm{spa} &= [[S_{{i,1}}^{1},...,S_{{i,{P^2}}}^{1}],...,[S_{{i,1}}^{D},...,S_{{i,{P^2}}}^{D}]]
\end{aligned}
\end{equation}
where ${S_i}^\mathrm{spe} \in \mathbb{R}^{1\times D \times P^2}$ and ${S_i}^\mathrm{spa} \in \mathbb{R}^{1\times P^2 \times D}$. This dual-domain strategy effectively captures spatial-spectral contextual features through inherent interaction modeling. 

The obtained 1D sequences require channel grouping for processing in the multi-group Mamba, with domain-specific implementations across both spatial and spectral domains to preserve cross-modal interactions, mathematically expressed as
\begin{equation}
\begin{aligned}
{\overline{S}_i}^\mathrm{spe} &= \operatorname{split}({S_i}^\mathrm{spe}), \\
{\overline{S}_i}^\mathrm{spa} &= \operatorname{split}({S_i}^\mathrm{spa})
\end{aligned}
\end{equation}
where ${\overline{S}_i}^\mathrm{spe} \in \mathbb{R}^{1\times D\times M\times U}$ and ${\overline{S}_i}^\mathrm{spa} \in \mathbb{R}^{1\times P^2\times N\times V}$, M and N denotes the number of groups of spe and spa scanning modules respectively. $M\times U$ is equal to $P^2$ and $N\times V$ is equal to $D$.

\subsubsection{Multi-Groups Mamba}
HSI classification as a pixel-level high-dimensional classification task, requires effective learning of inter-pixel spatial relationships, inter-band spectral correlations, and their intrinsic interactions. This demands robust long-range dependency modeling capabilities in feature construction layers. Our framework addresses this challenge through multi-group Mamba modules as core components, which employ dynamic weight integration across grouped features to efficiently model long-range dependencies with linear computational complexity.

\begin{figure}[h]
\centering
\includegraphics[width=3.4in, keepaspectratio]{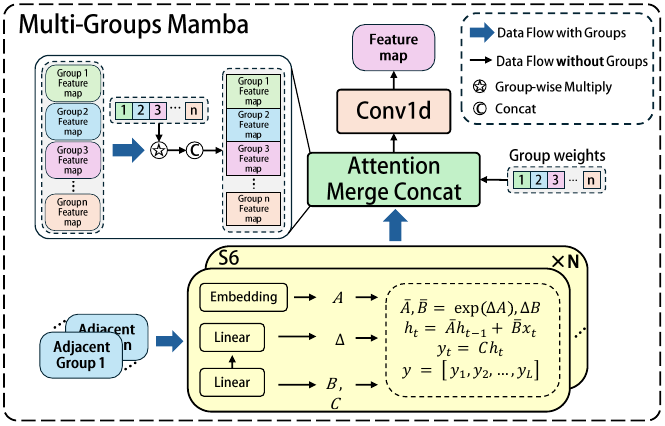}\\
\caption{The proposed multi-groups Mamba module employs parallel S6\cite{Mamba} as sub-modules to independently process grouped input features, learn their long-range dependencies, and fuse outputs through element-wise multiplication with learnable weights, followed by channel-wise concatenation to reconstruct original dimensions.}
\label{fig:multigroups-Mamba}
\end{figure}

The specific architecture is shown in Fig. \ref{fig:multigroups-Mamba}. Given the input ${\overline{S}_i} \in \mathbb{R}^{1\times \mathrm{S_L}\times \mathrm{N_G}\times \mathrm{D_G}}$ and the weight list corresponding to the group length $w = [w_1,w_2,...,w_\mathrm{N_G}]$, where $\mathrm{S_L}$ denotes the sequence length, $\mathrm{N_G}$ and $\mathrm{D_G}$ denotes the group count and the single-group channel length, mathematically expressed as
\begin{equation}
\begin{aligned}
F_\mathrm{mamba} &= \operatorname{Concat}\left( [ w_i \odot \operatorname{S6}_i(\overline{S}_i) ]_{i=1}^\mathrm{N_G} \right)_{\mathrm{channel}}
\end{aligned}
\end{equation}
where $F_{mamba} \in \mathbb{R}^{1\times \mathrm{S_L}\times (N_G\times D_G)}$. This grouped adaptive fusion mechanism effectively captures intra-group dependencies while mitigating computational overhead in long-range modeling, with enhanced focus on region-specific core features within localized group partitions.

\subsubsection{Multi-scale Patch Extraction}
\begin{figure}[h]
\centering
\includegraphics[width=2.6in, keepaspectratio]{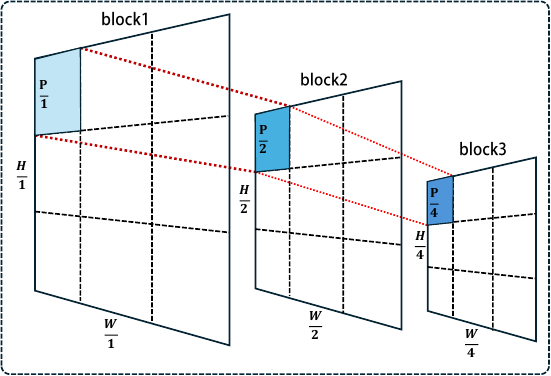}\\
\caption{An illustrative diagram demonstrating feature map evolution and adaptive patch size variation in the DCSS-Encoder, revealing its hierarchical multi-scale feature learning process.}
\label{fig:patch}
\end{figure}

This hierarchical approach progressively captures multi-scale features while adaptively adjusting receptive fields through recursively halving both the feature map dimensions $(\frac{H}{2^{i-1}},\frac{W}{2^{i-1}})$ and patch sizes $\frac{P}{2^{i-1}}$, where $i \in [1, 2, 3]$ denotes three HS-Mamba Blocks. The specific architecture is shown in Fig. \ref{fig:patch}. By synchronously scaling spatial granularity and feature resolution, it eliminates redundant computations in non-overlapping patch scanning, maintains structural consistency across scales, and optimizes computational efficiency without sacrificing multi-level feature representation capability.

\subsection{Lightweight Global Inline Attention}
The standalone DCSS-Encoder inadequately captures global context when processing patch tokens. We therefore introduce two lightweight global attention mechanisms, Spe-compressed Atten and Spa-extended Atten, applied to dual domains. These mechanisms enhance holistic feature awareness by modeling full spectral band correlations and global spatial patterns. The specific architecture is shown in Fig. \ref{fig:block}.

\subsubsection{Spe-compressed Atten}
The spe-compressed part combines dual-pooling and bottleneck convolutions to compress spectral features, learning channel dependencies via spatial aggregation while balancing computational efficiency and discriminative capacity, mathematically expressed as
\begin{equation}
\begin{aligned}
A_{\mathrm{avg}} &= \operatorname{AvgPool}(F_\mathrm{image}),
A_{\mathrm{max}} = \operatorname{MaxPool}(F_\mathrm{image}), \\
A_{\mathrm{cat}} &= \operatorname{Concat}(A_{\mathrm{avg}}, A_{\mathrm{max}})_{\mathrm{channel}}, \\
W_\mathrm{spe} &=  \operatorname{Conv^{2}_{1\times 1}} \left( \operatorname{SiLU}\left( \operatorname{Conv^{1}_{1\times 1}}(A_{\mathrm{cat}}) \right) \right)
\end{aligned}
\end{equation}
where $A_{\mathrm{avg}} \in \mathbb{R}^{1\times \mathrm{2C}\times 1\times 1}$ denotes the concat output, $\operatorname{Conv^{1}_{1\times 1}}$ is used for dimension adjustment $\mathbb{R}^{2D} \to \mathbb{R}^{D/\tau}$ and $\operatorname{Conv^{2}_{1\times 1}}$ is used for dimension adjustment $\mathbb{R}^{D/\tau} \to \mathbb{R}^{D}$, $\tau$ is 4 in the paper.

\subsubsection{Spa-extended Atten}
The spa-extended part integrates spectral pooling and dilated convolution to generate spatial attention, aggregating cross-channel features and modeling long-range dependencies for parameter-efficient global context enhancement, mathematically expressed as
\begin{equation}
\begin{aligned}
A_{\mathrm{avg}} &= \frac{1}{D}\sum_{\mathrm{c}=1}^D F_\mathrm{image}, \quad 
A_{\mathrm{max}} = \max_\mathrm{channel} F_\mathrm{image}, \\
A_{\mathrm{cat}} &= \operatorname{Concat}(A_{\mathrm{avg}}, A_{\mathrm{max}})_{\mathrm{channel}}, \\
W_\mathrm{spa} &= \text{SiLU}\left( \mathop{\operatorname{Conv_{3\times 3}}}\limits_{\substack{ \mathrm{dilation}=2}} (A_{\mathrm{cat}}) \right)
\end{aligned}
\end{equation}
where $A_{\mathrm{avg}} \in \mathbb{R}^{1\times \mathrm{2}\times \mathrm{H}\times \mathrm{W}}$ denotes the concat output, $\operatorname{Conv_{3\times 3}}$ with extend dilation is used for dimension adjustment $\mathbb{R}^{2} \to \mathbb{R}^{1}$.

\section{Experiments}
In this section, we first introduce four public HSI datasets for comparison. Then, we introduce the experimental settings, including evaluation metrics, comparison methods and implement details. Finally, we conduct detailed quantitative comparative experiment and ablation study to evaluate the effect of our proposed method.

\subsection{Datasets Description}
In the comparative experiments of this paper, we use four well-known public datasets: Indian Pines, Pavia University, WHU-Hi-HanChuan and WHU-Hi-HongHu. The division details for training, validation and test sets are provided in. The specific results are shown in Table \ref{tab:dataset_all}.
\subsubsection{Indian Pines}
The dataset was captured by the Airborne/Visible Infrared Imaging Spectrometer (AVIRIS) imaging an Indian pine tree over Northwestern Indiana in 1992. The imaging wavelength of the spectrometer ranges from 0.4 to 2.5 \textmu m. The dataset consists of corrected 200 spectral bands and 145 × 145 pixels, with a spatial resolution of 20 m per pixel. There are 10249 ground sample points, representing 16 distinct types such as Alfalfa, Corn-notill, and Corn-mintill.
\subsubsection{Pavia University}
The dataset was acquired by the Reflective Optics System Imaging Spectrometer (ROSIS) over the University of Pavia, Italy, this dataset covers a wavelength range of 0.43–0.86 \textmu m with 103 spectral bands. The image size is 610 × 340 pixels at a 1.3 m spatial resolution, containing 42776 labeled samples categorized into 9 classes such as Asphalt, Meadows, and Bare Soil.
\subsubsection{WHU-Hi-HanChuan}
The dataset was captured by the HiWHU sensor over Hanchuan, Hubei Province, China, this dataset spans 400–1000 nm with 274 spectral bands. It features a high spatial resolution of 0.463 m and an image size of 1217 × 303 pixels. The dataset includes 123797 annotated samples across 16 agricultural and urban land cover categories, including Rice, Cotton, and Plastic Greenhouses.
\subsubsection{WHU-Hi-HongHu}
The dataset was collected via the HSI-600 system over Honghu Lake, Hubei Province, this dataset covers 400–1000 nm with 270 spectral bands. It provides 0.6 m spatial resolution imagery (940 × 475 pixels) and contains approximately 120,000 labeled samples spanning 22 wetland vegetation and water body types, such as Reed, Lotus, and Open Water.

\begin{table*}
\centering
\caption{Abbreviated Category Colors, Names and Sample Numbers on Four Public Datasets}
\label{tab:dataset_all}
\begin{tabular}{
    c|
    cp{0.3cm}p{0.3cm}p{0.7cm}|
    cp{0.3cm}p{0.3cm}p{0.7cm}|
    cp{0.3cm}p{0.3cm}p{0.7cm}|
    cp{0.3cm}p{0.3cm}p{0.7cm}
}
\hline
\hline
\multirow{2}{*}{No.}&
\multicolumn{4}{c|}{Indian Pines}&
\multicolumn{4}{c|}{Pavia University}&
\multicolumn{4}{c|}{WHU-Hi-HanChuan}&
\multicolumn{4}{c}{WHU-Hi-HongHu}\\\cline{2-17}
& Cat. & Train & Val & Test 
& Cat. & Train & Val & Test 
& Cat. & Train & Val & Test 
& Cat. & Train & Val & Test\\
\hline
\cellcolor{MyColor1} 1 & Alfalfa & 30 & 10 & 6 & Asphalt & 30 & 10 & 6591 & Strawb. & 30 & 10 & 44695 & Red-r. & 30 & 10 & 14001 \\
\cellcolor{MyColor2} 2 & Corn-nt. & 30 & 10 & 1388 & Meadows & 30 & 10 & 18609 & Cowpea & 30 & 10 & 22713 & Road & 30 & 10 & 3472 \\
\cellcolor{MyColor3} 3 & Corn-mt. & 30 & 10 & 790 & Gravel & 30 & 10 & 2059 & Soybean & 30 & 10 & 10247 & Bare-s. & 30 & 10 & 21781 \\
\cellcolor{MyColor4} 4 & Corn & 30 & 10 & 197 & Trees & 30 & 10 & 3024 & Sorghum & 30 & 10 & 5313 & Cotton & 30 & 10 & 163245 \\
\cellcolor{MyColor5} 5 & Grass-p. & 30 & 10 & 443 & Metal-sh. & 30 & 10 & 1305 & Water-sp. & 30 & 10 & 1160 & Cotton-fw. & 30 & 10 & 6178 \\
\cellcolor{MyColor6} 6 & Grass-t. & 30 & 10 & 690 & Bare-s. & 30 & 10 & 4989 & Waterm. & 30 & 10 & 4493 & Rape & 30 & 10 & 44517 \\
\cellcolor{MyColor7} 7 & Grass-pm. & 15 & 5 & 8 & Bitumen & 30 & 10 & 1290 & Greens & 30 & 10 & 5863 & Ch-cabbage & 30 & 10 & 24063 \\
\cellcolor{MyColor8} 8 & Hay-w. & 30 & 10 & 438 & Bricks & 30 & 10 & 3642 & Trees & 30 & 10 & 17938 & Pakchoi & 30 & 10 & 4014 \\
\cellcolor{MyColor9} 9 & Oats & 10 & 5 & 5 & Shadows & 30 & 10 & 907 & Grass & 30 & 10 & 9429 & Cabbage & 30 & 10 & 10779 \\
\cellcolor{MyColor10} 10 & Soy-nt. & 30 & 10 & 932 &  &  &  &  & Red-r. & 30 & 10 & 10476 & Tuber-m. & 30 & 10 & 12354 \\
\cellcolor{MyColor11} 11 & Soy-mt. & 30 & 10 & 2415 &  &  &  &  & Gray-r. & 30 & 10 & 16871 & Bras-par. & 30 & 10 & 10975 \\
\cellcolor{MyColor12} 12 & Soy-cl. & 30 & 10 & 553 &  &  &  &  & Plastic & 30 & 10 & 3639 & Bras-ch. & 30 & 10 & 8914 \\
\cellcolor{MyColor13} 13 & Wheat & 30 & 10 & 165 &  &  &  &  & Bare-s. & 30 & 10 & 9076 & Sm-Bras-ch. & 30 & 10 & 22467 \\
\cellcolor{MyColor14} 14 & Woods & 30 & 10 & 1225 &  &  &  &  & Road & 30 & 10 & 18520 & Lactuca-s. & 30 & 10 & 7316 \\
\cellcolor{MyColor15} 15 & Bldgs. & 30 & 10 & 346 &  &  &  &  & Bright-obj. & 30 & 10 & 1096 & Celtuce & 30 & 10 & 962 \\
\cellcolor{MyColor16} 16 & Stone & 30 & 10 & 53 &  &  &  &  & Water & 30 & 10 & 75361 & Film-cl. & 30 & 10 & 7222 \\
\cellcolor{MyColor17} 17 &  &  &  &  &  &  &  &  &  &  &  &  & Romaine-l. & 30 & 10 & 2970 \\
\cellcolor{MyColor18} 18 &  &  &  &  &  &  &  &  &  &  &  &  & Carrot & 30 & 10 & 3177 \\
\cellcolor{MyColor19} 19 &  &  &  &  &  &  &  &  &  &  &  &  & White-rad. & 30 & 10 & 8672 \\
\cellcolor{MyColor20} 20 &  &  &  &  &  &  &  &  &  &  &  &  & Garlic-spr. & 30 & 10 & 3446 \\
\cellcolor{MyColor21} 21 &  &  &  &  &  &  &  &  &  &  &  &  & Broad-b. & 30 & 10 & 1288 \\
\cellcolor{MyColor22} 22 &  &  &  &  &  &  &  &  &  &  &  &  & Tree & 30 & 10 & 4000 \\
\hline
 & total & 445 & 150 & 9654 & total & 270 & 90 & 42416 & total & 480 & 160 & 256890 & total & 660 & 220 & 385813\\
\hline
\hline
\end{tabular}
\end{table*}

\subsection{Experimental Settings}
\subsubsection{Evaluation Metrics}
The experiments employed three widely-used evaluation metrics in HSI classification: Overall Accuracy (OA), Average Accuracy (AA), and Kappa coefficient (Kappa). To mitigate inherent random deviations, all key comparative and ablation studies were conducted with fixed random seeds for 10 consecutive runs, with final results reported as the mean ± standard deviation.
\subsubsection{Comparison Methods}
We conducted comparative experiments with 8 state-of-the-art (SOTA) models to comprehensively assess the effect of our proposed method.
\begin{itemize}
\item \textbf{SVM}\cite{SVM}: The model takes pixel-patch HSI data as input and adopts SVM to complete HSI classification.
\item \textbf{3D-CNN}\cite{3D-CNN}: The model examines HSI data from a 3D perspective, takes pixel-patch HSI data as input and adopts 3D-CNN to complete HSI classification.
\item \textbf{FullyContNet}\cite{FullyContext-CNN}: The model takes the whole-image HSI data as input and combines the multi-scale attention mechanism to complete HSI classification.
\item \textbf{SSFTT}\cite{SSFTT-Transformer}: The model takes pixel-patch HSI data as input and mixes CNN and Transformer to extract features hierarchically to complete HSI classification.
\item \textbf{MorphFormer}\cite{MorphFormer}: The model takes pixel-patch HSI data as input, uses spectral and spatial morphological convolution, combined with self-attention mechanism, to complete HSI classification and get SOTA performance in TGRS2023.
\item \textbf{GSC-ViT}\cite{GSC-ViT}: The model takes pixel-patch HSI data as input and adopts grouped separable convolution ViT to capture local and global spectral-spatial information to complete HSI classification.
\item \textbf{3DSS-Mamba}\cite{3DSS-Mamba}: The model takes pixel-patch HSI data as input, also examines the 3D characteristics of HSI data, uses a 3D dual-domain scanning mechanism, and combines Mamba's powerful long-range modeling capabilities to complete HSI classification.
\item \textbf{MambaHSI}\cite{MambaHSI}: The model takes the whole image HSI data as input and is the first Mamba-based model to process the whole image. A spatial-spectral feature extractor is designed to fuse the dual-domain data in an adaptive way to complete HSI classification.
\end{itemize}

\subsubsection{Implement Details}
HS-Mamba is implemented in PyTorch with the following experimental setup: For each class, 30 pixel samples are allocated for training, 10 for validation, and the remaining for testing. The S6 block parameters (timescale $\Delta$, projection matrices $B$ and $C$) inherit configurations from Mamba\cite{Mamba}. Training employs the ${Adam}$ optimizer (initial learning rate 0.0003) with ${CrossEntropyLoss}$ for backpropagation.  The number of groups for each GN is set to 8, and the number of channels in the intermediate process is fixed to 128. All experiments are conducted on a system with NVIDIA Quadro RTX 8000 GPUs, Intel Xeon Gold 6254R CPU, 512GB RAM, and CUDA 12.2.

\begin{table*}[htbp]
\centering
\caption{Quantitative Result (ACC\% ± STD\%) of Indian Pines Dataset. The Best in \textbf{Bold} and the Second with \underline{Underline}.}
\label{tab:contrastIP}
\begin{tabular}{c|c|cc|ccc|ccc}
\hline  
\hline
\multirow{3}{*}{Class}& \multicolumn{1}{c|}{ML-based} & \multicolumn{2}{c|}{CNN-based} & \multicolumn{3}{c|}{Transformer-based} & \multicolumn{3}{c}{Mamba-based}\\\cline{2-10}
& SVM & 3D-CNN & FullyContNet & SSFTT & MorphFormer & GSC-ViT & 3DSS-Mamba & MambaHSI & HS-Mamba\\
& {\tiny TGRS2004} & {\tiny TGRS2020} & {\tiny TGRS2022} & {\tiny TGRS2022} & {\tiny TGRS2023} & {\tiny TGRS2024} & {\tiny TGRS2024} & {\tiny TGRS2024} & {\tiny ours}\\
\hline
1 & 93.12±5.19 & \textbf{100} & \textbf{100} & \textbf{100} & \textbf{100} & \textbf{100} & \underline{96.67±6.67} & \textbf{100} & \textbf{100} \\
2 & 53.55±4.05 & 44.15±9.12 & \textbf{94.21±5.43} & 79.60±4.67 & 77.26±9.93 & 83.30±4.06 & 77.06±7.47 & 89.17±4.09 & \underline{89.61±4.00} \\
3 & 59.60±6.17 & 57.06±7.15 & \underline{92.90±3.59} & 87.87±3.25 & 90.20±4.91 & 93.34±4.12 & 84.39±5.37 & 90.80±2.23 & \textbf{93.19±2.68} \\
4 & 71.88±6.76 & 88.38±3.42 & \textbf{99.85±0.33} & 98.12±1.32 & 98.93±1.43 & 98.83±1.56 & 96.60±2.13 & 97.72±3.09 & \underline{98.93±1.64} \\
5 & 86.69±1.78 & 77.13±8.44 & 92.53±3.59 & \textbf{93.86±1.81} & 93.36±2.67 & 92.84±3.40 & 90.86±3.10 & \underline{93.66±2.27} & 93.23±2.51 \\
6 & 89.56±4.23& 92.35±2.25 & \underline{98.20±1.31} & 97.81±1.90 & 97.65±2.53 & 97.43±2.87 & 95.91±2.48 & 98.13±0.89 & \textbf{98.87±1.04} \\
7 & 89.23±10.43 & 97.50±5.00 & \textbf{100} & \textbf{100} & \textbf{100} & \textbf{100} & \textbf{100} & \underline{98.75±3.75} & \textbf{100} \\
8 & 90.71±2.99 & 95.14±3.03 & \textbf{100} & 98.42±2.03 & 99.75±0.35 & \textbf{100} & 98.13±1.92 & \underline{99.98±0.07} & \textbf{100} \\
9 & 88.00±15.36 & \textbf{100} & \textbf{100} & \textbf{100} & \textbf{100} & \textbf{100} & \underline{94.00±12.81} & \textbf{100} & \textbf{100} \\
10 & 66.74±4.24 & 63.87±6.22 & 89.82±2.92 & 83.71±4.43 & 85.91±3.95 & 89.54±3.93 & 79.51±7.70 & \underline{92.39±3.16} & \textbf{93.53±2.53} \\
11 & 53.46±4.16 & 57.76±8.48 & \underline{88.35±4.78} & 70.63±9.17 & 77.47±5.49 & 78.53±10.13 & 73.23±5.54 & 88.29±2.52 & \textbf{93.36±3.22} \\
12 & 60.75±4.73 & 62.69±7.00 & 88.30±2.72 & 90.36±5.15 & 90.94±4.42 & \textbf{92.66±2.93} & 83.20±8.71 & 91.79±2.64 & \underline{92.08±2.48} \\
13 & 96.97±1.98 & 99.27±0.71 & \underline{99.94±0.18} & 99.64±0.40 & 99.70±0.41 & \textbf{100} & 99.09±1.19 & 99.94±0.18 & \textbf{100} \\
14 & 82.92±5.96 & 82.24±8.31 & 98.13±1.48 & 95.46±2.52 & 95.80±2.03 & 91.84±5.68 & 92.72±3.67 & \underline{98.4±0.81} & \textbf{99.04±0.71} \\
15 & 52.87±4.42 & 81.45±4.06 & 98.13±1.86 & 92.60±5.39 & 95.03±3.41 & 98.01±2.20 & 93.84±3.84 & \underline{99.54±0.74} & \textbf{99.57±1.12} \\
16 & 96.35±2.36 & 99.43±0.86 & 99.06±1.52 & \textbf{99.81±0.57} & 98.68±2.39 & \underline{99.43±1.21} & 96.98±2.82 & 98.68±1.90 & 98.30±2.45 \\
\hline
OA(\%) & 66.76±1.50 & 67.27±3.06 & 91.82±1.10 & 85.21±3.19 & 87.19±2.85 & 88.61±2.07 & 83.74±2.24 & \underline{92.85±0.81} & \textbf{94.65±0.91} \\
AA(\%) & 77.02±1.38 & 81.15±2.02 & 95.64±0.48 & 92.99±1.40 & 93.79±1.65 & 94.73±0.76 & 90.76±1.60 & \underline{96.08±0.51} & \textbf{96.86±0.55} \\
Kappa(\%) & 62.55±1.69 & 63.04±3.33 & 90.66±1.24 & 83.23±3.53 & 85.43±3.23 & 87.05±2.31 & 81.53±2.51 & \underline{91.82±0.93} & \textbf{93.87±1.04} \\
\hline
\hline
\end{tabular}
\end{table*}

\begin{table*}[htbp]
\centering
\caption{Quantitative Result (ACC\% ± STD\%) of Pavia University Dataset. The Best in \textbf{Bold} and the Second with \underline{Underline}.}
\label{tab:contrastPU}
\vspace{-0.5em} % 向上压缩
\begin{tabular}{c|c|cc|ccc|ccc}
\hline 
\hline
\multirow{3}{*}{Class}& \multicolumn{1}{c|}{ML-based} & \multicolumn{2}{c|}{CNN-based} & \multicolumn{3}{c|}{Transformer-based} & \multicolumn{3}{c}{Mamba-based}\\\cline{2-10}
& SVM & 3D-CNN & FullyContNet & SSFTT & MorphFormer & GSC-ViT & 3DSS-Mamba & MambaHSI & HS-Mamba\\
& {\tiny TGRS2004} & {\tiny TGRS2020} & {\tiny TGRS2022} & {\tiny TGRS2022} & {\tiny TGRS2023} & {\tiny TGRS2024} & {\tiny TGRS2024} & {\tiny TGRS2024} & {\tiny ours}\\
\hline
1 & 73.63±3.86 & 74.27±5.51 & 83.14±3.84 & 86.68±2.37 & 90.35±5.01 & 92.54±3.71 & 83.75±4.99 & \textbf{94.37±1.89} & \underline{94.19±4.32} \\
2 & 76.61±5.84 & 71.20±4.37 & 92.17±3.41 & 90.48±6.89 & 93.90±3.86 & 93.10±3.32 & 88.35±3.95 & \underline{96.05±1.76} & \textbf{96.55±2.77} \\
3 & 77.22±6.74 & 67.51±7.20 & \underline{95.76±2.71} & 82.59±4.87 & 84.21±5.88 & 88.27±5.34 & 84.95±6.84 & 93.38±5.16 & \textbf{99.05±1.08} \\
4 & 92.23±3.95 & 94.79±2.06 & 94.51±2.64 & \textbf{96.68±2.79} & 94.83±2.18 & \underline{94.98±3.46} & 91.14±3.36 & 88.28±3.87 & 89.97±3.17 \\
5 & 99.14±0.29 & 99.66±0.37 & 99.89±0.14 & \underline{99.90±0.16} & 99.79±0.42 & 99.74±0.48 & 99.49±0.97 & \textbf{99.98±0.05} & 99.89±0.23 \\
6 & 78.13±8.13 & 75.51±9.64 & 98.58±2.01 & 92.39±4.52 & 95.87±3.36 & 95.16±3.86 & 90.82±6.43 & \underline{98.64±0.98} & \textbf{99.85±0.25} \\
7 & 91.87±1.66 & 85.60±6.48 & \underline{98.79±1.08} & 93.40±2.03 & 94.75±2.28 & 97.81±3.08 & 95.29±2.43 & 96.72±2.87 & \textbf{99.96±0.08} \\
8 & 81.18±4.67 & 82.28±4.61 & 89.58±3.12 & 89.41±4.69 & 91.55±3.28 & 95.01±2.08 & 83.90±8.97 & \underline{94.29±2.76} & \textbf{96.00±1.55} \\
9 & \textbf{99.85±0.05} & 99.48±0.43 & 95.53±1.16 & \underline{99.51±0.42} & 98.81±1.65 & 99.27±0.64 & 98.03±1.48 & 99.45±0.62 & 98.79±0.75 \\
\hline
OA(\%) & 79.53±2.01 & 76.56±1.11 & 92.22±1.79 & 90.65±3.16 & 93.28±1.75 & 93.80±1.62 & 88.34±2.72 & \underline{95.47±0.84} & \textbf{96.43±1.35} \\
AA(\%) & 85.54±1.22& 83.37±1.34 & 94.55±0.84 & 92.34±1.26 & 93.78±1.00 & 95.10±1.33 & 90.64±1.83 & \underline{95.68±1.11} & \textbf{97.14±0.81} \\
Kappa(\%) & 73.83±2.34 & 70.26±1.28 & 89.83±2.25 & 87.84±3.94 & 91.18±2.25 & 91.86±2.09 & 84.88±3.40 & \underline{94.02±1.10} & \textbf{95.29±1.75} \\
\hline
\hline
\end{tabular}
\end{table*}

\begin{table*}[htbp]
\centering
\caption{Quantitative Result (ACC\% ± STD\%) of HanChuan Dataset. The Best in \textbf{Bold} and the Second with \underline{Underline}.}
\label{tab:contrastHC}
\vspace{-0.5em} % 向上压缩
\begin{tabular}{c|c|cc|ccc|ccc}
\hline  
\hline
\multirow{3}{*}{Class}& \multicolumn{1}{c|}{ML-based} & \multicolumn{2}{c|}{CNN-based} & \multicolumn{3}{c|}{Transformer-based} & \multicolumn{3}{c}{Mamba-based}\\\cline{2-10}
& SVM & 3D-CNN & FullyContNet & SSFTT & MorphFormer & GSC-ViT & 3DSS-Mamba & MambaHSI & HS-Mamba\\
& {\tiny TGRS2004} & {\tiny TGRS2020} & {\tiny TGRS2022} & {\tiny TGRS2022} & {\tiny TGRS2023} & {\tiny TGRS2024} & {\tiny TGRS2024} & {\tiny TGRS2024} & {\tiny ours}\\
\hline
1 & 68.09±3.22 & 68.72±5.65 & 91.47±3.31 & 75.68±8.65 & 85.53±5.00 & 87.30±3.40 & 74.05±5.51 & \underline{94.92±2.20} & \textbf{96.87±2.00} \\
2 & 36.01±5.74 & 63.13±9.99 & 77.35±4.84 & 74.14±8.65 & 78.28±10.63 & 74.42±7.33 & 64.57±7.90 & \underline{82.24±5.79} & \textbf{89.85±5.94} \\
3 & 65.53±4.91 & 64.83±7.82 & 90.75±5.52 & 74.46±7.36 & 85.53±9.19 & 87.01±8.52 & 80.23±8.35 & \underline{96.25±1.87} & \textbf{96.61±3.01} \\
4 & 89.77±2.77 & 89.76±3.69 & \underline{98.86±1.30} & 91.80±4.19 & 97.33±1.58 & 91.19±1.34 & 95.80±1.63 & 98.83±1.32 & \textbf{98.97±0.98} \\
5 & 68.50±5.76 & 93.63±3.61 & \textbf{100} & 96.45±3.06 & 98.22±1.42 & 97.50±1.12 & 94.33±3.35 & \underline{99.76±0.31} & 99.69±0.28 \\
6 & 35.26±5.62 & 31.15±7.45 & \underline{88.33±4.42} & 57.32±5.95 & 73.04±7.64 & 69.89±10.87 & 61.69±9.62 & 88.10±5.53 & \textbf{94.58±3.64} \\
7 & 84.61±3.37 & 94.50±2.19 & \underline{97.61±1.79} & 89.91±4.61 & 90.47±7.43 & 86.54±5.89 & 87.53±5.35 & 97.36±1.61 & \textbf{99.66±0.45} \\
8 & 43.38±4.41 & 57.18±2.37 & \underline{87.47±3.16} & 65.91±6.24 & 69.48±3.90 & 64.68±6.43 & 68.39±4.32 & 83.16±4.59 & \textbf{94.87±2.83} \\
9 & 35.97±5.57 & 31.79±5.62 & 90.79±2.47 & 72.51±7.26 & 76.64±5.26 & 73.86±7.82 & 68.40±5.27 & \underline{91.25±3.70} & \textbf{94.38±2.60} \\
10 & 83.56±2.82 & 81.12±5.17 & \underline{95.92±3.48} & 88.87±6.27 & 95.41±3.10 & 96.66±1.26 & 91.18±3.76 & 94.62±2.86 & \textbf{97.10±2.05} \\
11 & 92.99±3.20 & 87.55±9.39 & \underline{97.18±1.81} & 81.98±9.57 & 91.22±7.64 & 87.65±5.62 & 87.13±5.38 & 92.90±4.24 & \textbf{97.65±1.23} \\
12 & 55.04±5.39 & 56.47±6.69 & \textbf{99.38±1.38} & 82.86±7.62 & 89.29±5.14 & 86.54±4.18 & 87.24±5.52 & 95.97±3.73 & \underline{99.32±0.88} \\
13 & 51.55±5.19 & 39.75±8.32 & 79.59±3.20 & 62.27±5.53 & 69.29±9.32 & 75.11±7.65 & 63.12±8.51 & \underline{84.53±4.14} & \textbf{89.12±3.30} \\
14 & 62.97±4.57 & 63.30±4.32 & \underline{94.71±1.20} & 72.26±8.25 & 78.32±5.11 & 76.43±7.24 & 75.14±5.01 & 88.44±4.45 & \textbf{96.16±2.24} \\
15 & 72.00±2.17 & 90.99±3.19 & 94.12±1.20 & \textbf{97.24±1.13} & 95.82±5.11 & 94.83±2.77 & 90.78±3.80 & 94.32±2.41 & \underline{94.68±2.16} \\
16 & 81.14±4.18 & 77.97±6.43 & 96.37±1.57 & 94.15±2.28 & 95.04±4.64 & \underline{97.94±2.00} & 95.51±4.18 & 97.69±1.03 & \textbf{98.54±0.60} \\
\hline
OA(\%) & 67.47±1.64 & 69.35±2.96 & 92.10±0.70 & 80.68±2.46 & 86.21±2.66 & 86.32±1.93 & 81.29±2.35 & \underline{92.78±1.08} & \textbf{96.64±0.79} \\
AA(\%) & 64.15±0.86 & 68.24±1.82 & 92.49±0.59 & 79.61±2.12 & 85.56±1.68 & 84.53±1.63 & 80.32±2.49 & \underline{92.52±0.89} & \textbf{96.13±0.67} \\
Kappa(\%) & 62.95±1.75 & 65.12±3.17 & 90.80±0.80 & 77.66±2.78 & 84.01±2.99 & 84.08±2.21 & 78.36±2.67 & \underline{91.58±1.25} & \textbf{95.72±0.92} \\
\hline
\hline
\end{tabular}
\end{table*}

\begin{table*}[htbp]
\renewcommand\arraystretch{1}
\centering
\caption{Quantitative Result (ACC\% ± STD\%) of HongHu Dataset. The Best in \textbf{Bold} and the Second with \underline{Underline}.}
\label{tab:contrastHH}
\vspace{-0.5em} % 向上压缩
\begin{tabular}{c|c|cc|ccc|ccc}
\hline
\hline
\multirow{3}{*}{Class}& \multicolumn{1}{c|}{ML-based} & \multicolumn{2}{c|}{CNN-based} & \multicolumn{3}{c|}{Transformer-based} & \multicolumn{3}{c}{Mamba-based}\\\cline{2-10}
& SVM & 3D-CNN & FullyContNet & SSFTT & MorphFormer & GSC-ViT & 3DSS-Mamba & MambaHSI & HS-Mamba\\
& {\tiny TGRS2004} & {\tiny TGRS2020} & {\tiny TGRS2022} & {\tiny TGRS2022} & {\tiny TGRS2023} & {\tiny TGRS2024} & {\tiny TGRS2024} & {\tiny TGRS2024} & {\tiny ours}\\
\hline
1 & 94.30±1.90 & 83.07±6.31 & \textbf{96.78±1.23} & 90.35±5.38 & 94.22±1.34 & 92.27±4.53 & 86.46±5.61 & 96.11±2.26 & \underline{96.28±2.12} \\
2 & 53.77±4.54 & 84.92±4.83 & \textbf{96.94±1.26} & 89.58±7.41 & 92.40±4.17 & 88.02±8.85 & 85.14±5.76 & 95.06±2.52 & \underline{95.56±3.06} \\
3 & 90.70±3.65 & 77.59±3.63 & 93.05±2.19 & 82.60±3.45 & 81.97±5.01 & 77.69±10.57 & 78.65±5.30 & \textbf{94.31±1.60} & \underline{93.20±2.18} \\
4 & \underline{97.42±0.34} & 83.57±3.58 & 86.97±3.40 & 89.51±2.75 & 93.78±2.53 & 93.66±3.66 & 84.30±4.19 & 95.73±2.47 & \textbf{98.57±1.13} \\
5 & 17.61±2.78 & 77.23±5.98 & \underline{98.94±0.96} & 86.71±4.52 & 91.28±3.09 & 87.66±5.17 & 86.34±4.09 & 96.11±2.90 & \textbf{99.14±0.74} \\
6 & 90.59±1.93 & 87.70±2.08 & 92.96±2.05 & 89.06±3.56 & 92.58±2.05 & 90.68±3.12 & 89.32±3.55 & \underline{92.96±3.62} & \textbf{97.43±1.06} \\
7 & 78.60±3.21 & 58.40±4.22 & 79.17±3.23 & 70.63±7.69 & 73.88±6.01 & 74.35±4.50 & 63.72±7.27 & \underline{86.60±3.55} & \textbf{88.88±1.25} \\
8 & 13.56±2.31 & 43.10±3.19 & \textbf{99.80±0.27} & 68.23±7.18 & 75.53±4.13 & 92.49±7.99 & 57.86±11.70 & 96.06±1.59 & \underline{99.36±0.74} \\
9 & \underline{96.57±1.71} & 93.79±1.15 & 95.40±1.70 & 95.85±1.69 & 96.46±1.93 & 96.29±1.47 & 93.47±2.18 & 94.50±2.91 & \textbf{97.26±1.55} \\
10 & 50.25±7.08 & 67.02±6.03 & 84.19±3.59 & 81.72±5.37 & 86.50±4.95 & 83.77±3.02 & 70.70±5.42 & \textbf{90.07±2.91} & \underline{88.67±2.88} \\
11 & 31.70±5.48 & 53.23±4.83 & 80.91±4.20 & 73.17±6.01 & 84.70±4.17 & 77.38±5.08 & 71.96±9.05 & \underline{90.51±3.36} & \textbf{92.82±3.93} \\
12 & 42.58±4.37 & 63.12±1.81 & \underline{88.30±6.59} & 76.29±4.49 & 77.83±5.94 & 76.30±7.65 & 67.10±6.75 & 87.05±5.80 & \textbf{92.12±3.59} \\
13 & 59.05±4.52 & 60.19±5.14 & 77.96±6.00 & 63.87±9.04 & 74.15±5.01 & 73.33±6.25 & 56.13±5.25 & \underline{88.51±2.51} & \textbf{89.72±2.85} \\
14 & 67.70±6.77 & 78.56±5.43 & 91.94±3.55 & 89.92±4.07 & \underline{94.43±2.33} & 88.66±4.86 & 81.83±5.87 & 93.90±1.46 & \textbf{95.59±1.67} \\
15 & 8.86±4.25 & 90.43±2.82 & 98.67±1.52 & 95.61±2.09 & 98.38±1.30 & 98.09±1.60 & 95.48±2.71 & \underline{98.79±1.03} & \textbf{99.41±0.62} \\
16 & 87.85±2.55 & 81.80±6.30 & 94.64±3.69 & 93.30±3.56 & 94.64±1.94 & 95.02±2.51 & 88.36±4.28 & \underline{96.83±1.78} & \textbf{98.72±1.21} \\
17 & 55.60±4.87 & 81.82±2.89 & \underline{99.00±2.41} & 91.24±4.81 & 92.60±5.93 & 89.36±11.90 & 86.11±7.61 & 98.16±2.39 & \textbf{99.74±0.40} \\
18 & 28.46±3.59 & 83.02±2.05 & \textbf{99.71±0.37} & 90.07±5.57 & 96.71±1.26 & 92.91±5.31 & 90.21±5.63 & 96.41±1.85 & \underline{98.49±1.17} \\
19 & 65.15±5.12 & 85.92±4.03 & 91.09±2.47 & 87.47±5.03 & 89.65±2.21 & 89.71±3.63 & 85.41±4.47 & \underline{91.32±2.87} & \textbf{92.56±3.14} \\
20 & 37.85±5.53 & 88.71±3.31 & \textbf{99.44±0.61} & 90.64±4.01 & 94.31±2.45 & 95.80±1.91 & 88.71±4.92 & 99.18±0.41 & \underline{98.82±0.95} \\
21 & 9.35±0.92 & 79.17±6.78 & \textbf{99.99±0.02} & 92.00±4.98 & 97.09±3.32 & 93.73±4.94 & 92.34±4.96 & \underline{99.82±0.26} & \textbf{99.99±0.02} \\
22 & 21.53±1.27 & 88.19±4.41 & \textbf{100} & 92.62±3.34 & 97.25±2.17 & 93.58±4.64 & 87.04±4.40 & 99.51±0.44 & \underline{99.72±0.68} \\
\hline
OA(\%) & 68.77±1.27 & 78.67±1.83 & 88.62±1.51 & 85.44±1.79 & 89.59±1.40 & 88.23±1.71 & 80.79±1.66 & \underline{93.80±1.42} & \textbf{96.10±0.74} \\
AA(\%) & 54.50±0.41 & 76.84±0.88 & 92.99±0.54 & 85.48±1.43 & 89.57±0.81 & 86.85±1.24 & 81.21±1.04 & \underline{94.43±0.76} & \textbf{96.00±0.53} \\
Kappa(\%) & 62.71±1.28 & 74.05±2.01 & 86.02±1.76 & 82.05±2.12 & 87.05±1.67 & 85.36±2.01 & 76.57±1.83 & \underline{92.24±1.74} & \textbf{95.08±0.92} \\
\hline
\hline
\end{tabular}
\end{table*}

\subsection{Comparison With SOTA Methods}
\subsubsection{Comparison on the IP Dataset}
\begin{figure*}[htbp]
\vspace{-1em} % 向上压缩
\centering
\begin{subfigure}{.12\textwidth}
\centering
\includegraphics[width=1\linewidth]{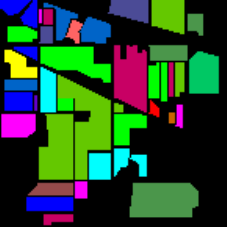}
\caption{}
\label{fig:gt_IP}
\end{subfigure}
\begin{subfigure}{.12\textwidth}
\centering
\includegraphics[width=1\linewidth]{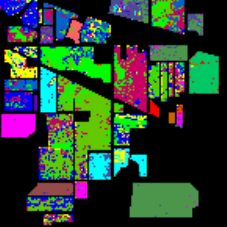}
\caption{}
\label{fig:SVM_IP}
\end{subfigure}
\begin{subfigure}{.12\textwidth}
\centering
\includegraphics[width=1\linewidth]{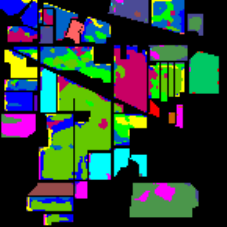}
\caption{}
\label{fig:CNN3d_IP}
\end{subfigure}
\begin{subfigure}{.12\textwidth}
\centering
\includegraphics[width=1\linewidth]{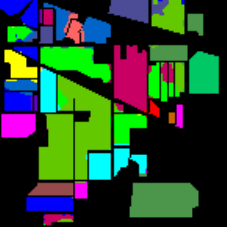}
\caption{}
\label{fig:Fully_IP}
\end{subfigure}
\begin{subfigure}{.12\textwidth}
\centering
\includegraphics[width=1\linewidth]{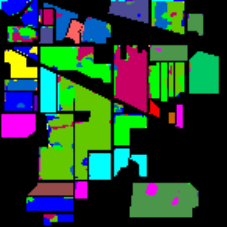}
\caption{}
\label{fig:SSFTT_IP}
\end{subfigure}
\\
\begin{subfigure}{.12\textwidth}
\centering
\includegraphics[width=1\linewidth]{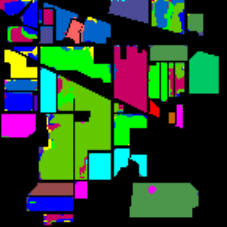}
\caption{}
\label{fig:MF_IP}
\end{subfigure}
\begin{subfigure}{.12\textwidth}
\centering
\includegraphics[width=1\linewidth]{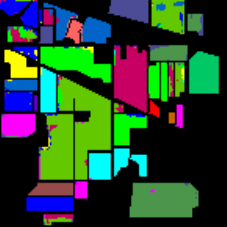}
\caption{}
\label{fig:GSC_IP}
\end{subfigure}
\begin{subfigure}{.12\textwidth}
\centering
\includegraphics[width=1\linewidth]{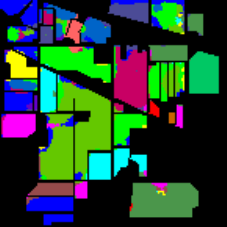}
\caption{}
\label{fig:TDSS_IP}
\end{subfigure}
\begin{subfigure}{.12\textwidth}
\centering
\includegraphics[width=1\linewidth]{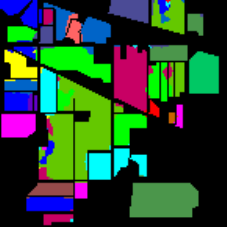}
\caption{}
\label{fig:MambaHSI_IP}
\end{subfigure}
\begin{subfigure}{.12\textwidth}
\centering
\includegraphics[width=1\linewidth]{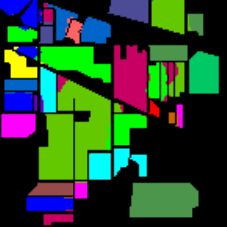}
\caption{}
\label{fig:proposed_IP}
\end{subfigure}
\caption{Visualization of the classification results for Indian Pines dataset. (a) Ground-truth map. (b) SVM. (c) 3D-CNN. (d) FullyContNet. (e) SSFTT. (f) MorpyFormer. (g) GSC-ViT. (h) 3DSS-Mamba. (i) MambaHSI. (j) proposed HS-Mamba. The meaning of colors refers to Table \ref{tab:dataset_all}.}
\label{fig:visualization_IP}
\vspace{-1em} % 向上压缩
\end{figure*}

As shown in Table \ref{tab:contrastIP}, our proposed HS-Mamba achieves the best performance on the most commonly used low-resolution HSI dataset IP. Specifically, HS-Mamba surpasses the second-best method MambaHSI by margins of 1.8\%, 0.78\%, and 2.05\% in OA, AA, and Kappa metrics, respectively. As shown in Fig. \ref{fig:visualization_IP}, in most categories such as Soybean-mintill, the classification is smoother and more accurate, demonstrating the robustness of our approach for low-resolution hyperspectral analysis.

\subsubsection{Comparison on the PU Dataset}
\begin{figure*}[htbp]
% \vspace{-1em} % 向上压缩
\centering
\begin{subfigure}{.12\textwidth}
\centering
\includegraphics[width=1\linewidth]{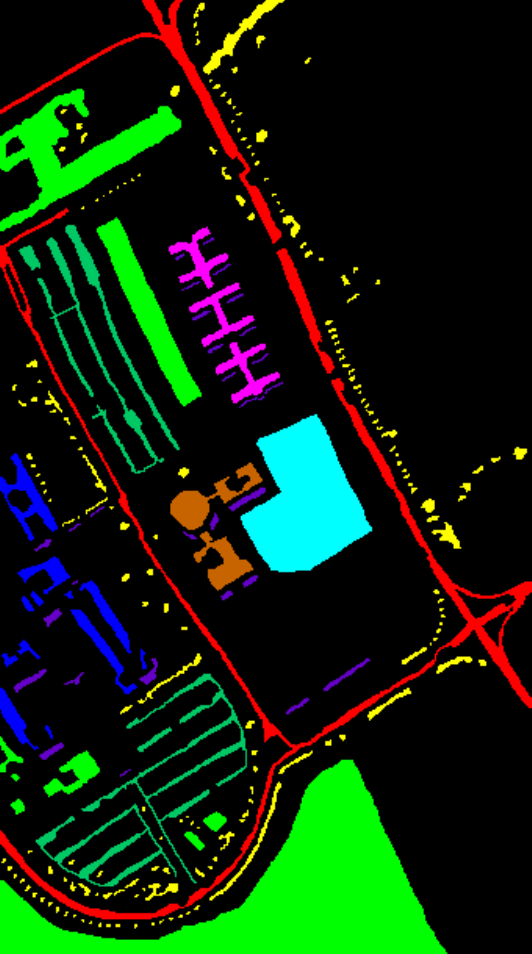}
\caption{}
\label{fig:gt_PU}
\end{subfigure}
\begin{subfigure}{.12\textwidth}
\centering
\includegraphics[width=1\linewidth]{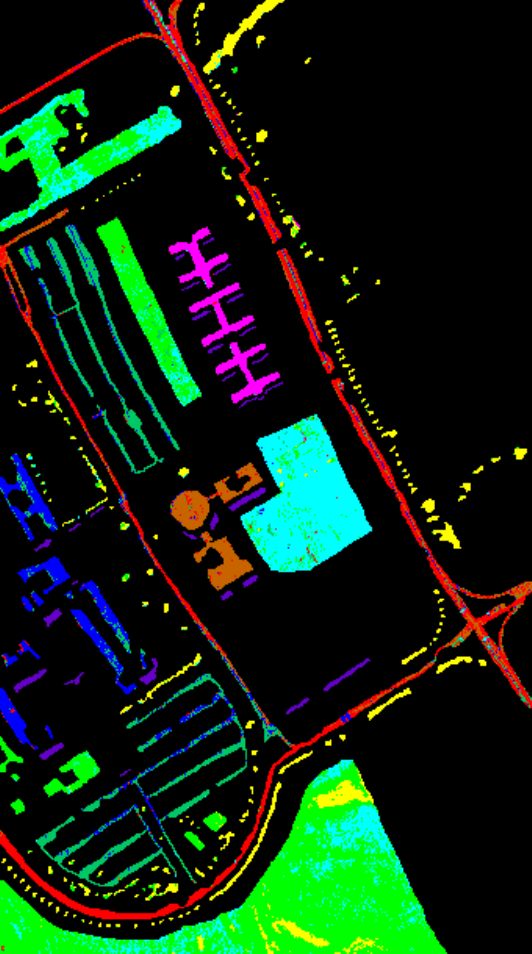}
\caption{}
\label{fig:SVM_PU}
\end{subfigure}
\begin{subfigure}{.12\textwidth}
\centering
\includegraphics[width=1\linewidth]{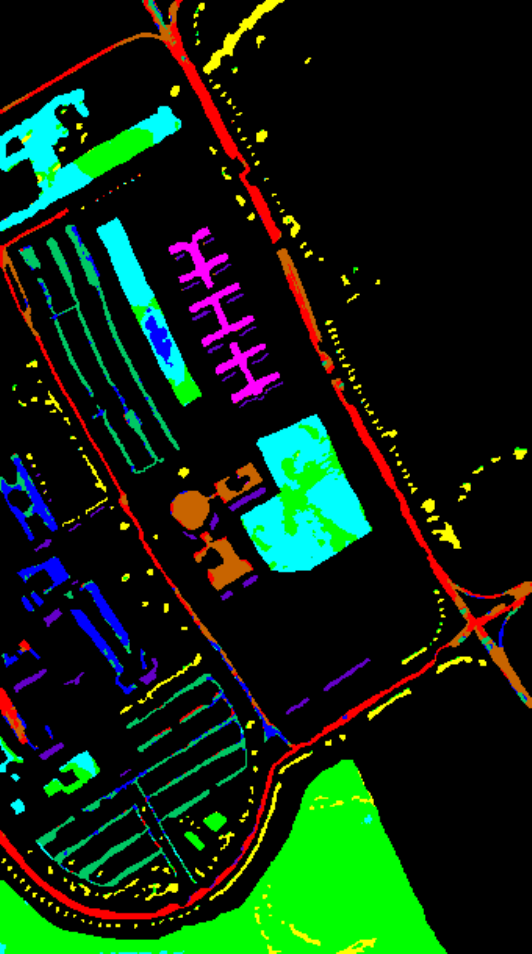}
\caption{}
\label{fig:CNN3d_PU}
\end{subfigure}
\begin{subfigure}{.12\textwidth}
\centering
\includegraphics[width=1\linewidth]{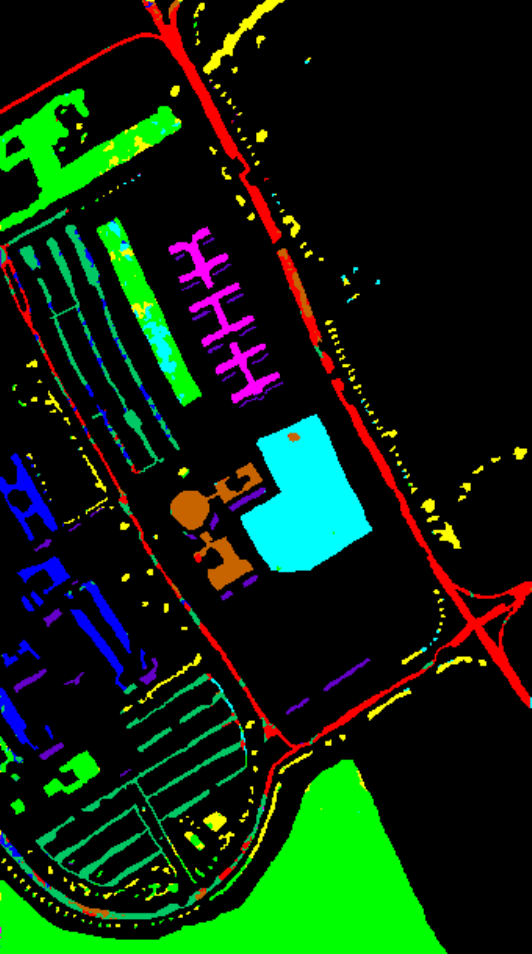}
\caption{}
\label{fig:Fully_PU}
\end{subfigure}
\begin{subfigure}{.12\textwidth}
\centering
\includegraphics[width=1\linewidth]{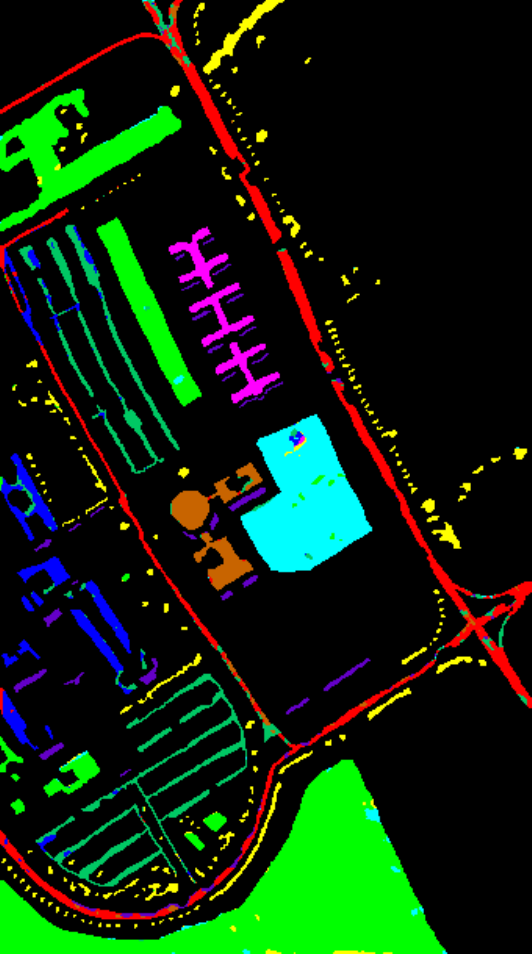}
\caption{}
\label{fig:SSFTT_PU}
\end{subfigure}
\\
\begin{subfigure}{.12\textwidth}
\centering
\includegraphics[width=1\linewidth]{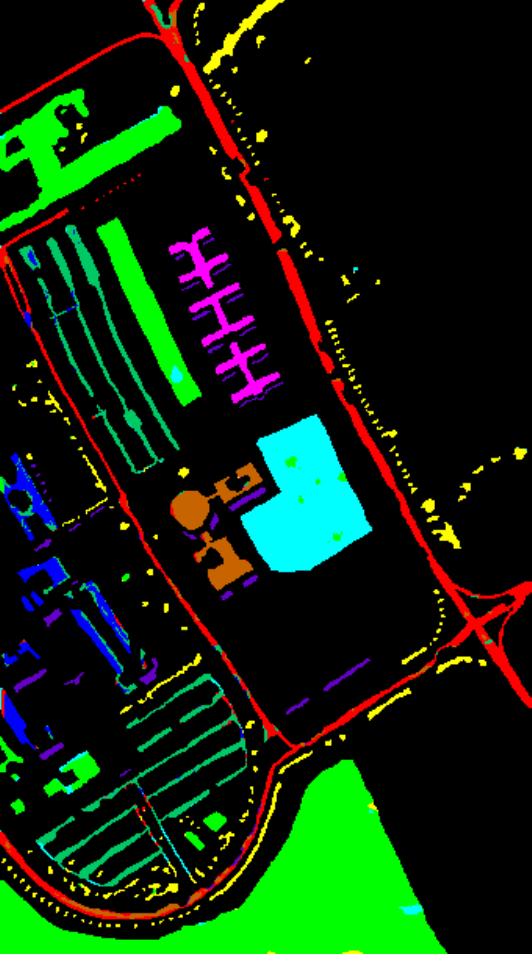}
\caption{}
\label{fig:MF_PU}
\end{subfigure}
\begin{subfigure}{.12\textwidth}
\centering
\includegraphics[width=1\linewidth]{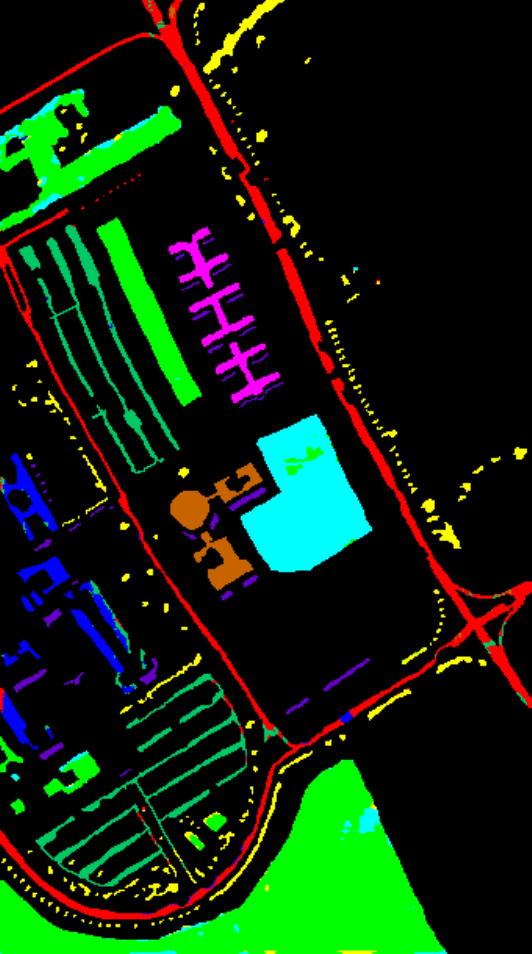}
\caption{}
\label{fig:GSC_PU}
\end{subfigure}
\begin{subfigure}{.12\textwidth}
\centering
\includegraphics[width=1\linewidth]{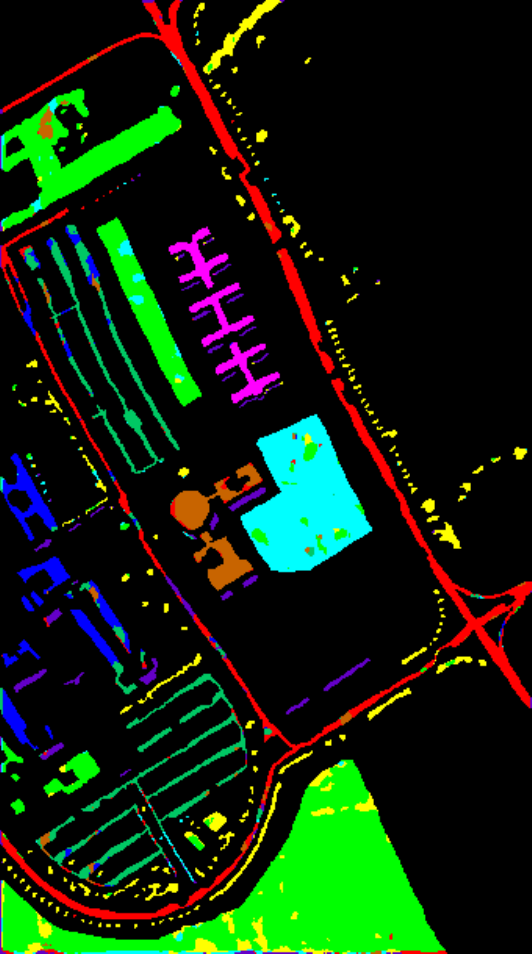}
\caption{}
\label{fig:TDSS_PU}
\end{subfigure}
\begin{subfigure}{.12\textwidth}
\centering
\includegraphics[width=1\linewidth]{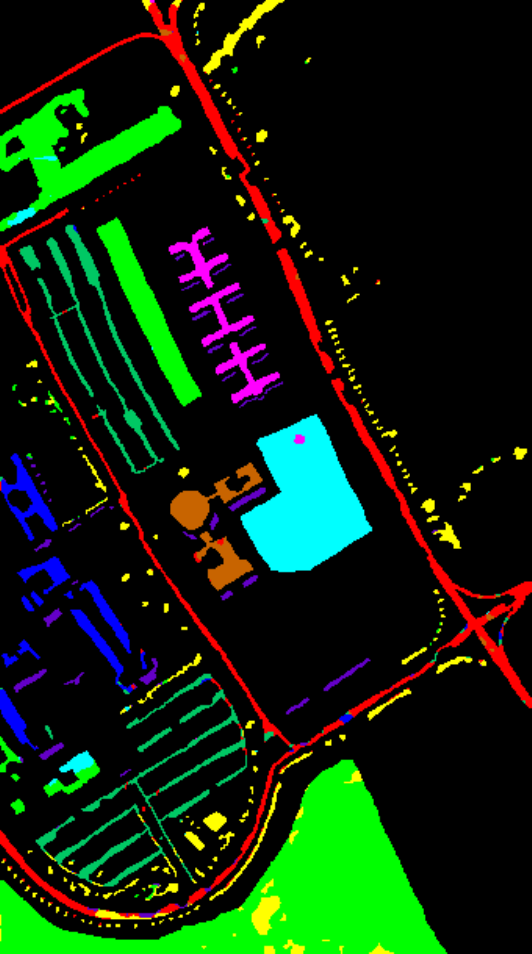}
\caption{}
\label{fig:MambaHSI_PU}
\end{subfigure}
\begin{subfigure}{.12\textwidth}
\centering
\includegraphics[width=1\linewidth]{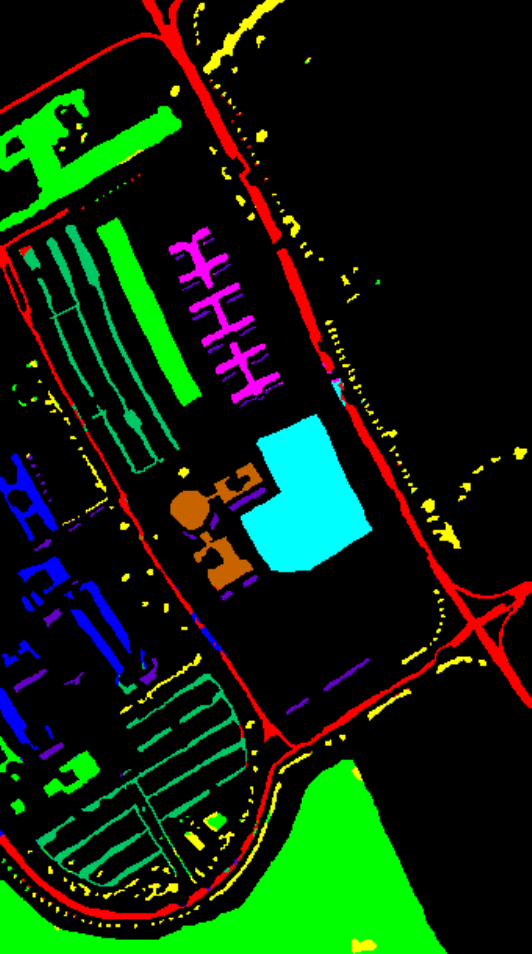}
\caption{}
\label{fig:proposed_PU}
\end{subfigure}
\caption{Visualization of the classification results for Pavia University dataset. (a) Ground-truth map. (b) SVM. (c) 3D-CNN. (d) FullyContNet. (e) SSFTT. (f) MorpyFormer. (g) GSC-ViT. (h) 3DSS-Mamba. (i) MambaHSI. (j) proposed HS-Mamba.}
\label{fig:visualization_PU}
\end{figure*}

As shown in Table \ref{tab:contrastPU}, our proposed HS-Mamba also achieves the best performance on the widely-used HSI medium-resolution dataset PU. Specifically, HS-Mamba surpasses the second-best method MambaHSI by margins of 0.96\%, 1.46\%, and 1.27\% in OA, AA, and Kappa metrics, respectively. As shown in Fig. \ref{fig:visualization_PU}, both Bare Soil and Meadows exhibit smoother classification boundaries, significantly reducing misclassified noise and delivering optimal results across most categories.

\subsubsection{Comparison on the HC Dataset}
\begin{figure*}[htbp]
\centering
\begin{subfigure}{.08\textwidth}
\centering
\includegraphics[width=1\linewidth]{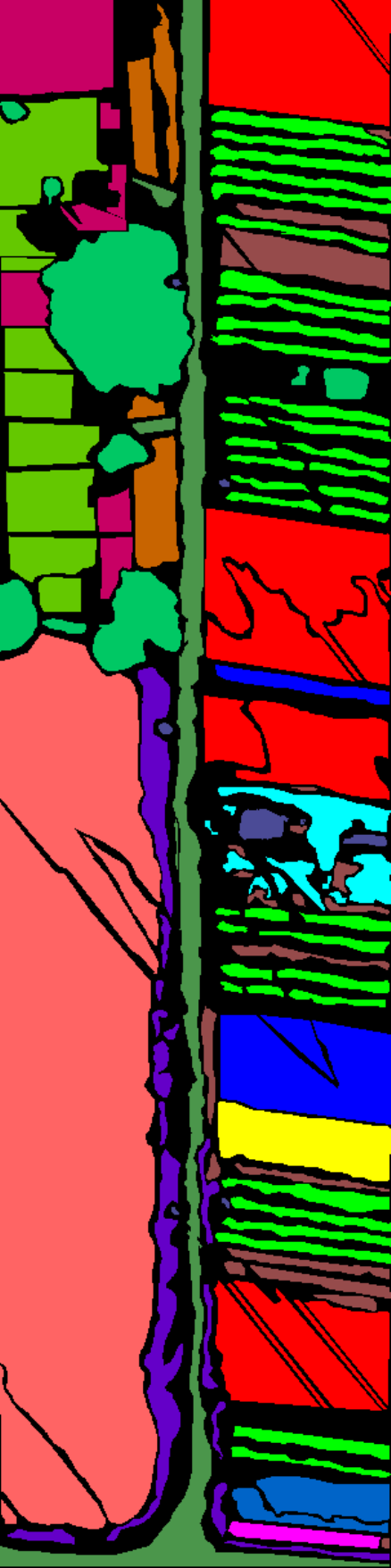}
\caption{}
\label{fig:gt_HC}
\end{subfigure}
\begin{subfigure}{.08\textwidth}
\centering
\includegraphics[width=1\linewidth]{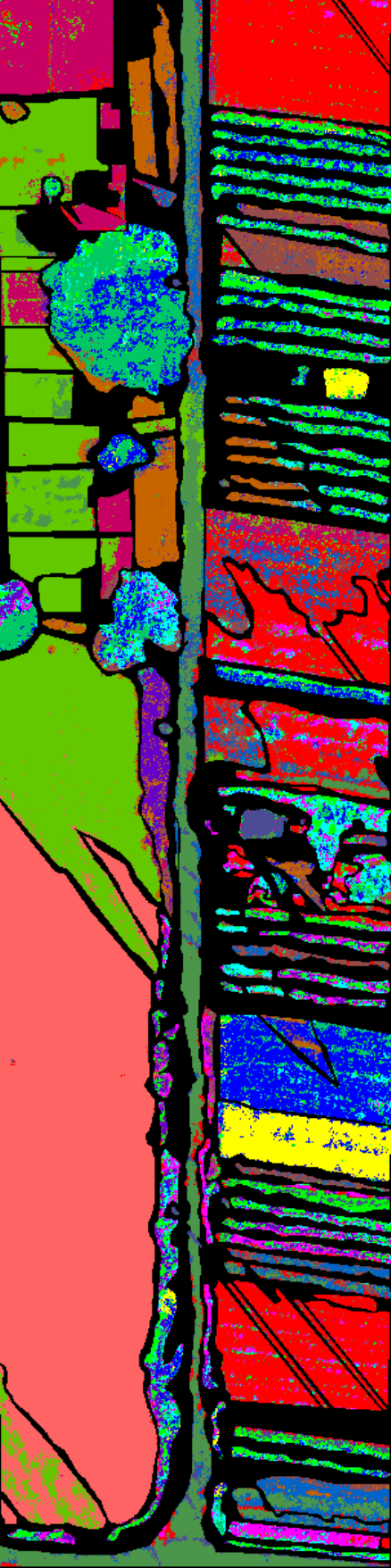}
\caption{}
\label{fig:SVM_HC}
\end{subfigure}
\begin{subfigure}{.08\textwidth}
\centering
\includegraphics[width=1\linewidth]{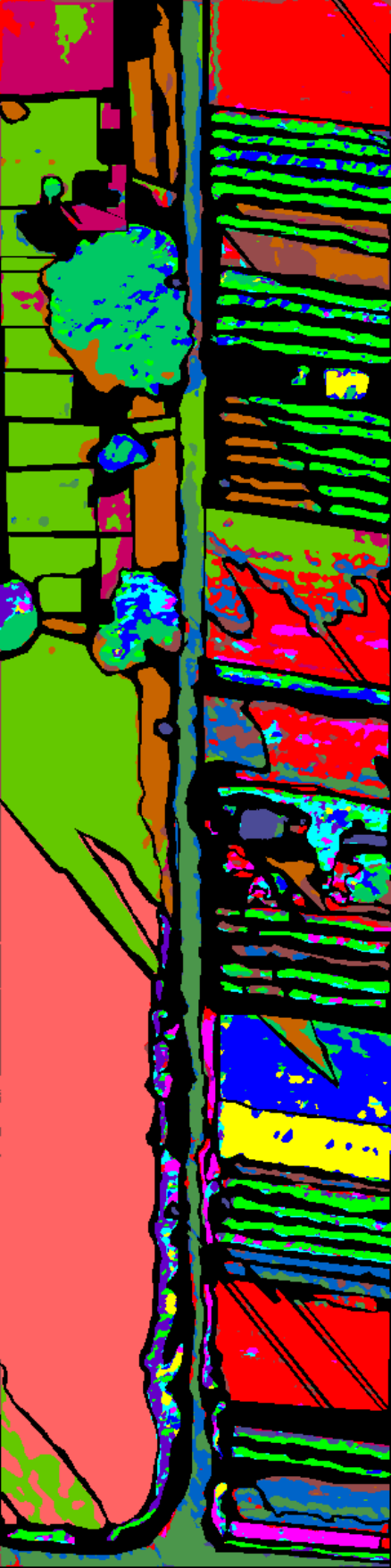}
\caption{}
\label{fig:CNN3d_HC}
\end{subfigure}
\begin{subfigure}{.08\textwidth}
\centering
\includegraphics[width=1\linewidth]{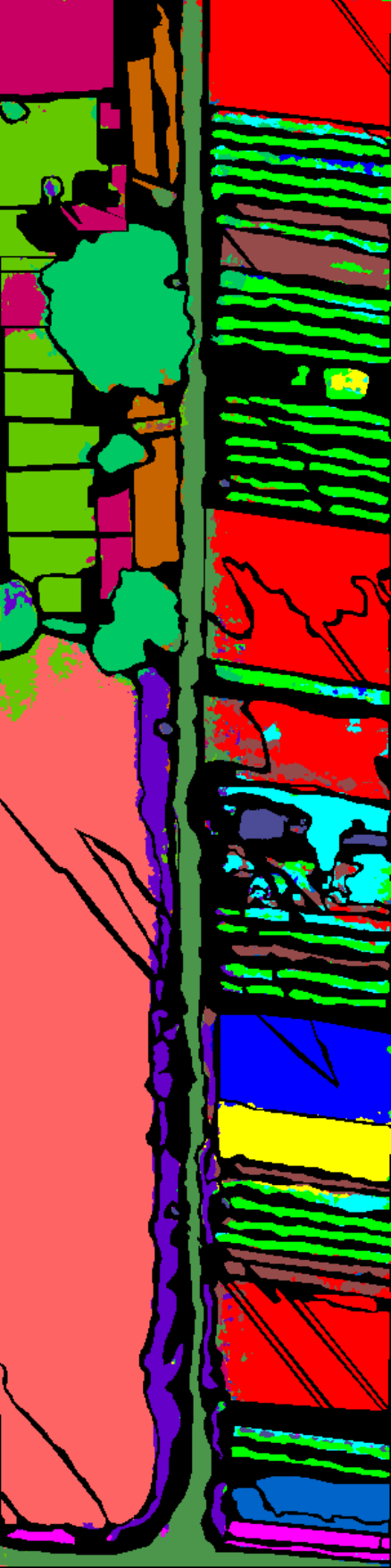}
\caption{}
\label{fig:Fully_HC}
\end{subfigure}
\begin{subfigure}{.08\textwidth}
\centering
\includegraphics[width=1\linewidth]{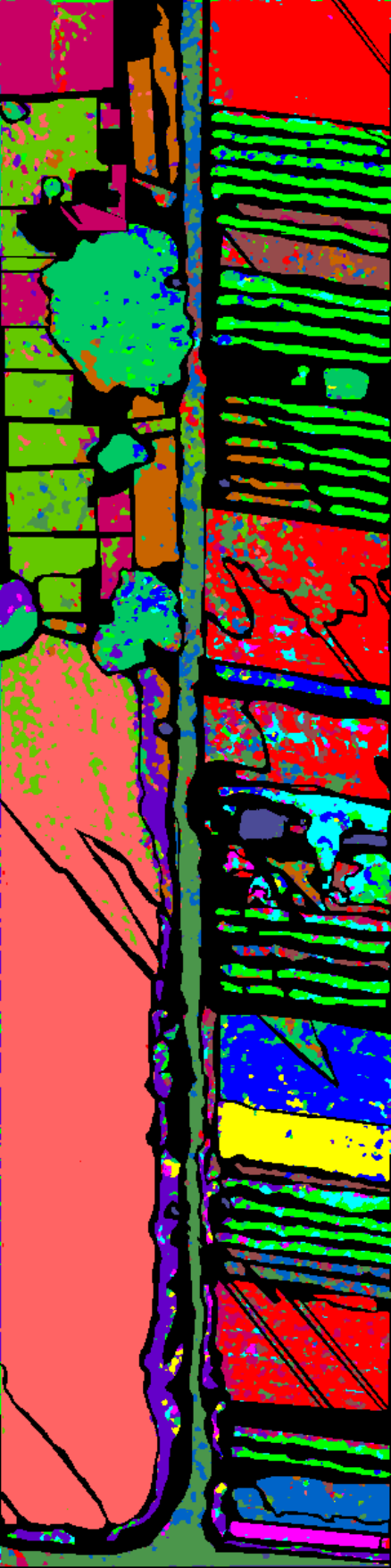}
\caption{}
\label{fig:SSFTT_HC}
\end{subfigure}
\begin{subfigure}{.08\textwidth}
\centering
\includegraphics[width=1\linewidth]{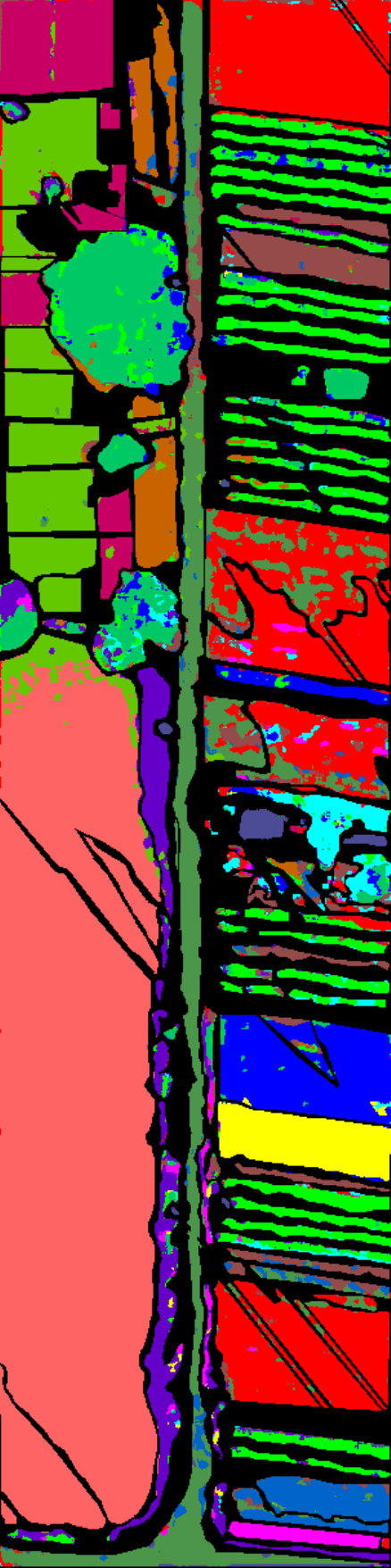}
\caption{}
\label{fig:MF_HC}
\end{subfigure}
\begin{subfigure}{.08\textwidth}
\centering
\includegraphics[width=1\linewidth]{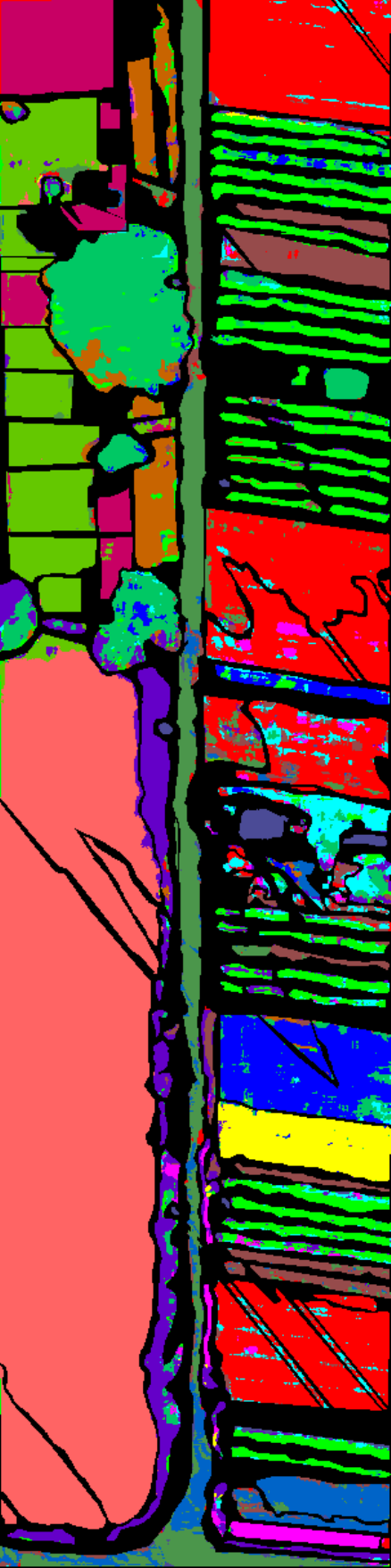}
\caption{}
\label{fig:GSC_HC}
\end{subfigure}
\begin{subfigure}{.08\textwidth}
\centering
\includegraphics[width=1\linewidth]{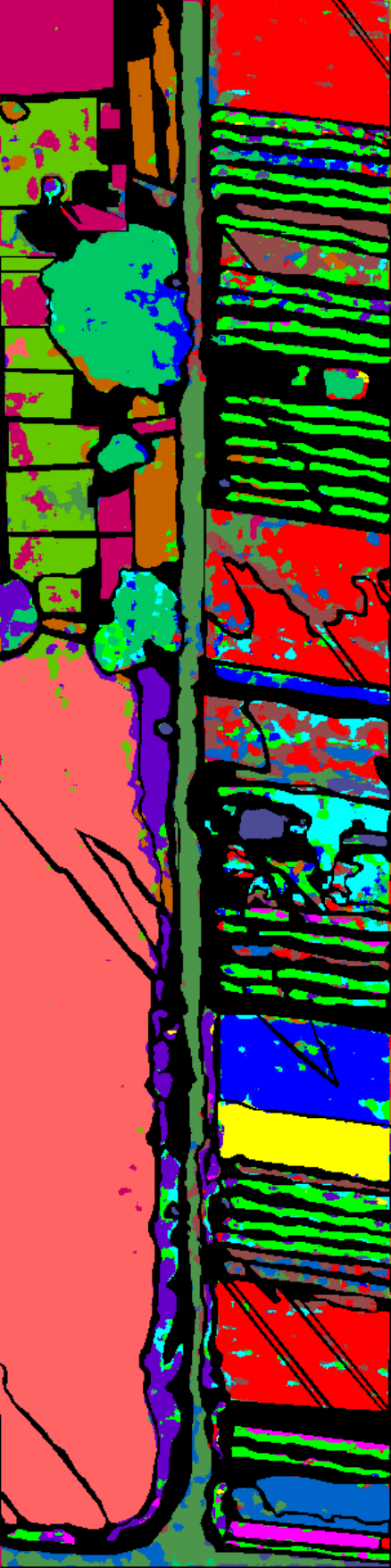}
\caption{}
\label{fig:TDSS_HC}
\end{subfigure}
\begin{subfigure}{.08\textwidth}
\centering
\includegraphics[width=1\linewidth]{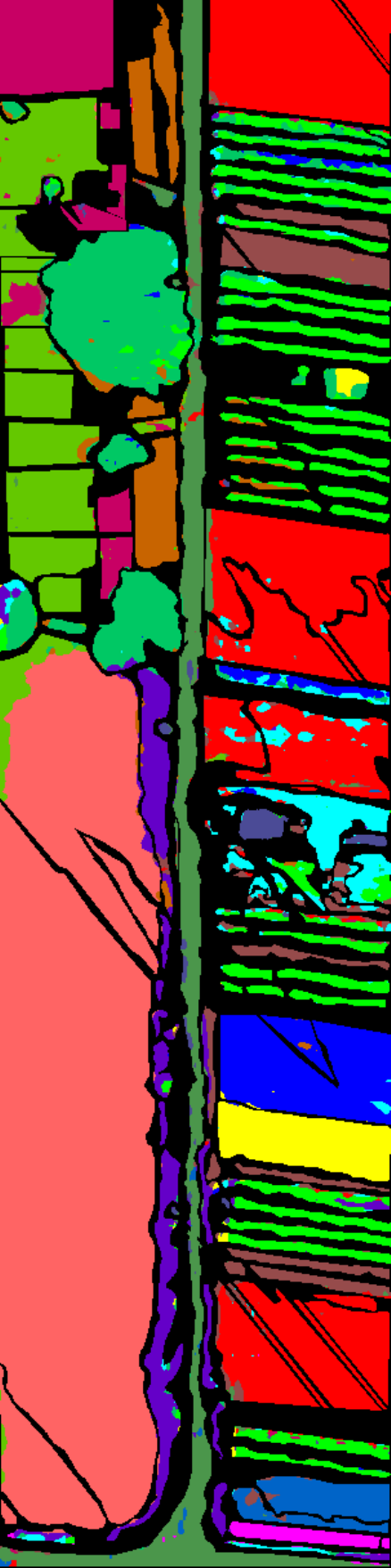}
\caption{}
\label{fig:MambaHSI_HC}
\end{subfigure}
\begin{subfigure}{.08\textwidth}
\centering
\includegraphics[width=1\linewidth]{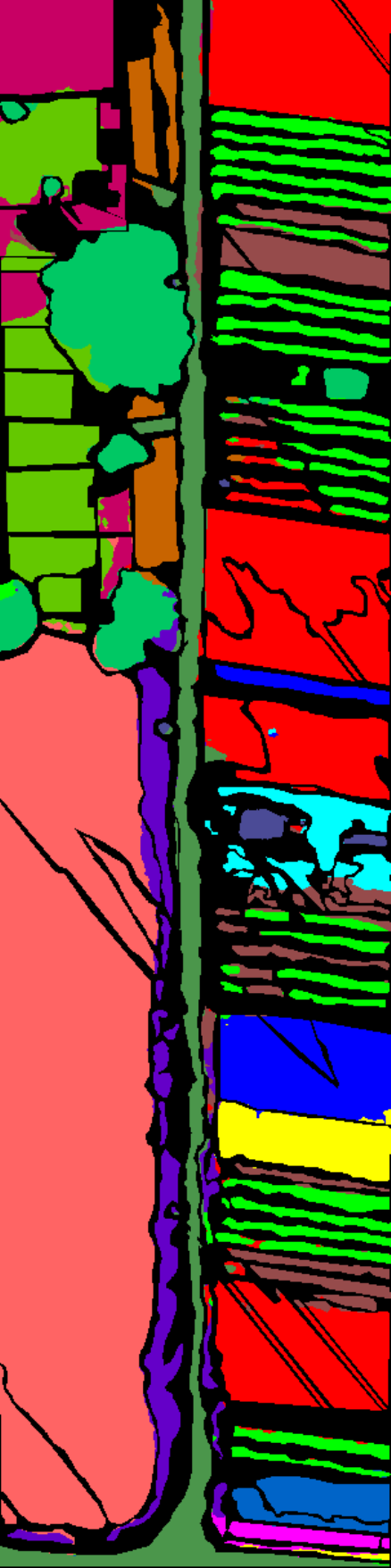}
\caption{}
\label{fig:proposed_HC}
\end{subfigure}
\caption{Visualization of the classification results for WHU-Hi-HanChuan dataset. (a) Ground-truth map. (b) SVM. (c) 3D-CNN. (d) FullyContNet. (e) SSFTT. (f) MorpyFormer. (g) GSC-ViT. (h) 3DSS-Mamba. (i) MambaHSI. (j) proposed HS-Mamba.}
\label{fig:visualization_HC}
\end{figure*}

As shown in Table \ref{tab:contrastHC}, our proposed HS-Mamba achieves superior performance on the wide-resolution HSI dataset. Specifically, HS-Mamba surpasses the second-best method MambaHSI by margins of 3.86\%, 3.61\%, and 4.14\% in OA, AA, and Kappa metrics, respectively. As shown in Fig. \ref{fig:visualization_HC}, common objects such as Trees and Water exhibit improved classification accuracy with significantly reduced errors compared to other methods, demonstrating the effect of our approach for wide-resolution HSI data.

\subsubsection{Comparison on the HH Dataset}
\begin{figure*}[htbp]
\centering
\begin{subfigure}{.12\textwidth}
\centering
\includegraphics[width=1\linewidth]{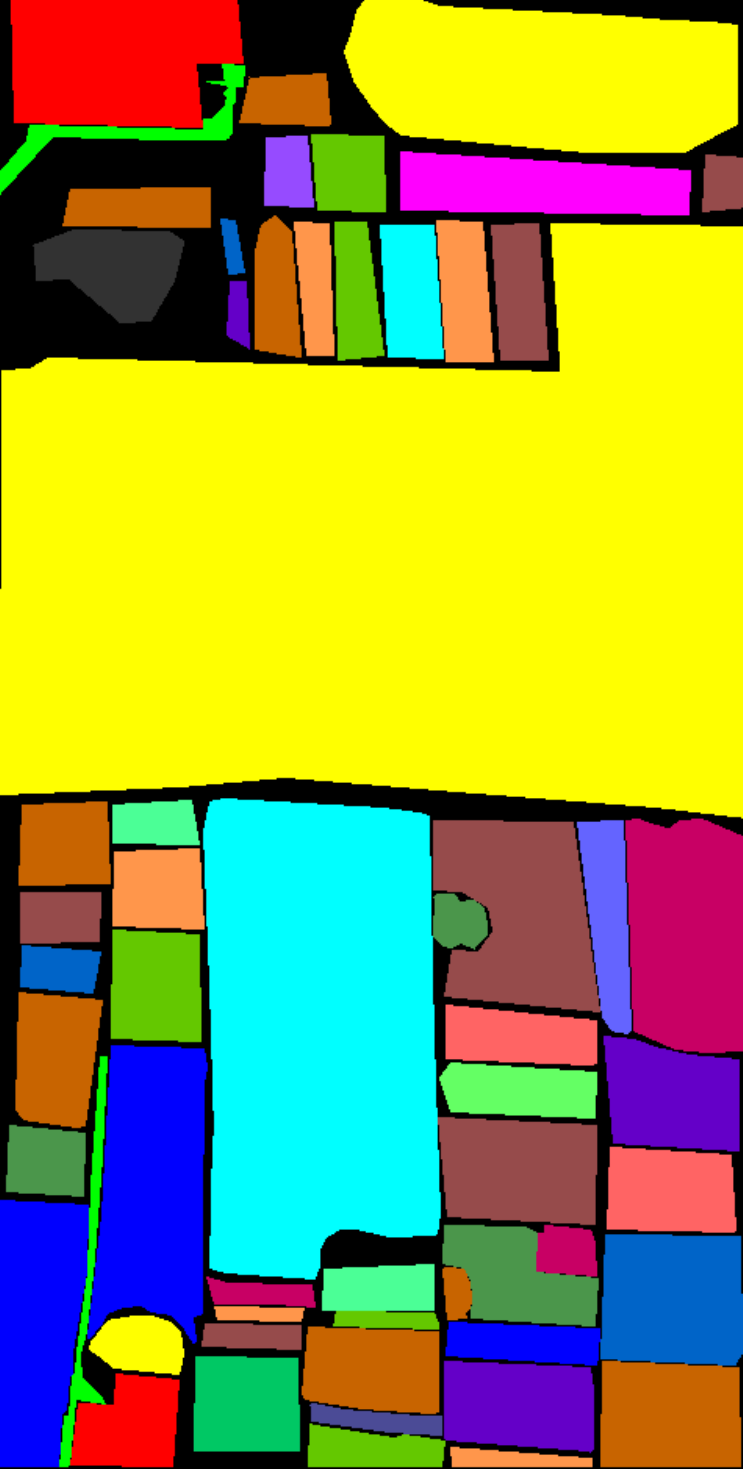}
\caption{}
\label{fig:gt_HH}
\end{subfigure}
\begin{subfigure}{.12\textwidth}
\centering
\includegraphics[width=1\linewidth]{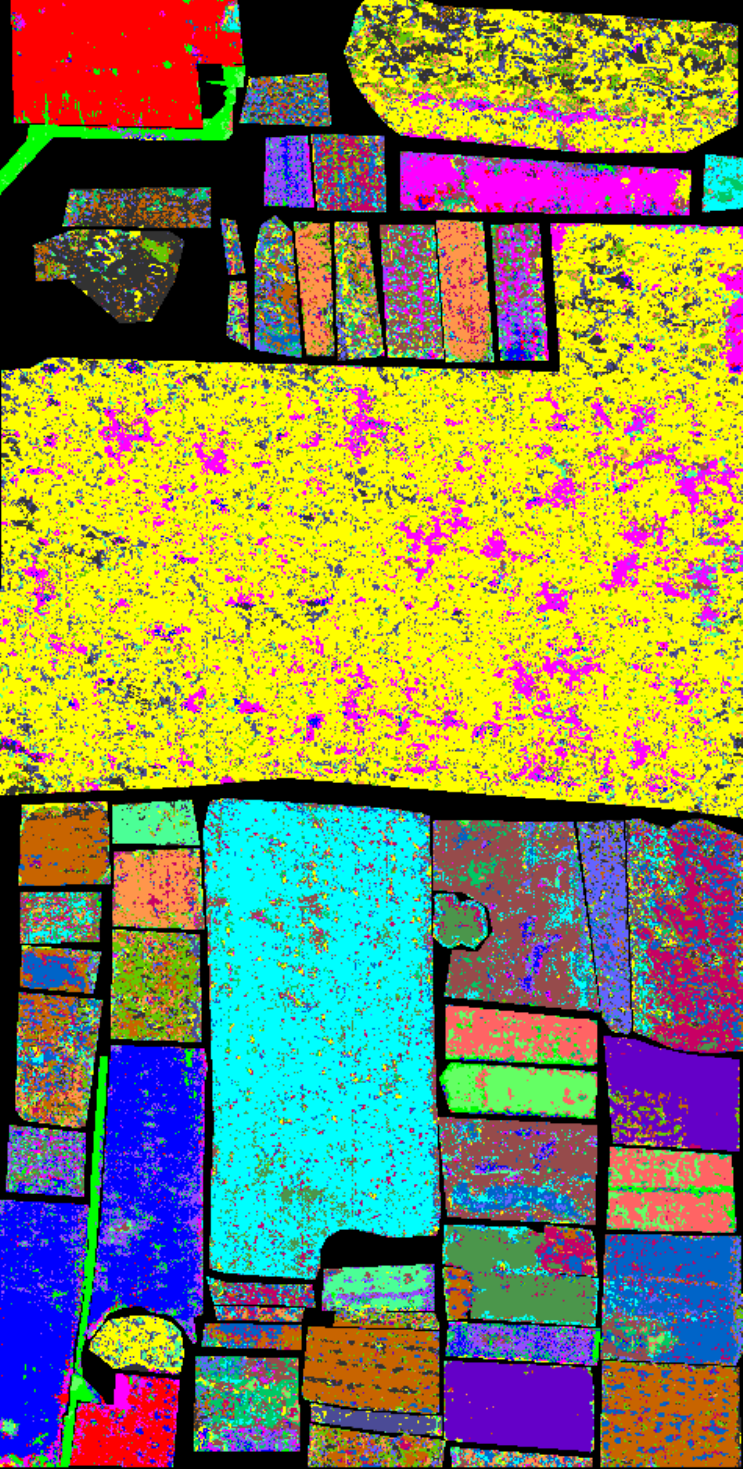}
\caption{}
\label{fig:SVM_HH}
\end{subfigure}
\begin{subfigure}{.12\textwidth}
\centering
\includegraphics[width=1\linewidth]{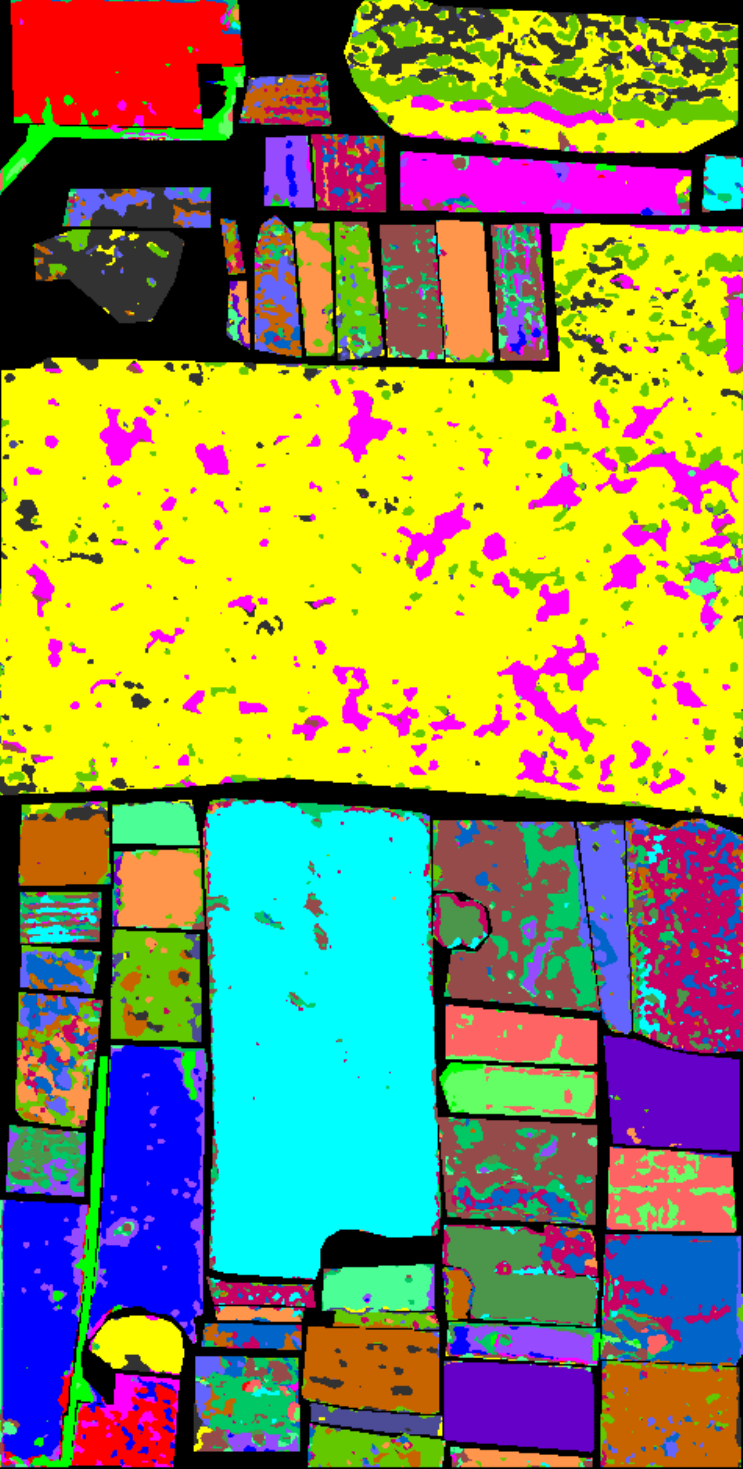}
\caption{}
\label{fig:CNN3d_HH}
\end{subfigure}
\begin{subfigure}{.12\textwidth}
\centering
\includegraphics[width=1\linewidth]{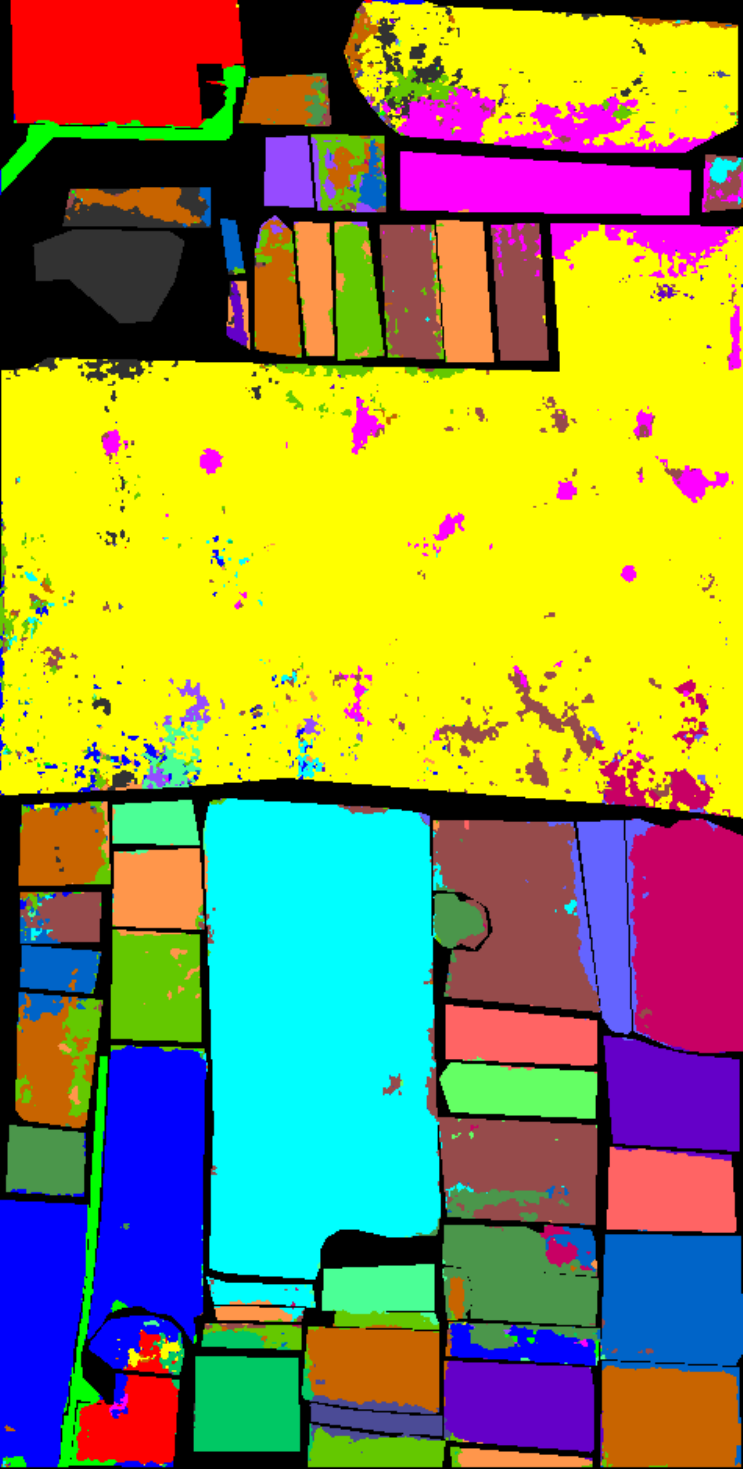}
\caption{}
\label{fig:Fully_HH}
\end{subfigure}
\begin{subfigure}{.12\textwidth}
\centering
\includegraphics[width=1\linewidth]{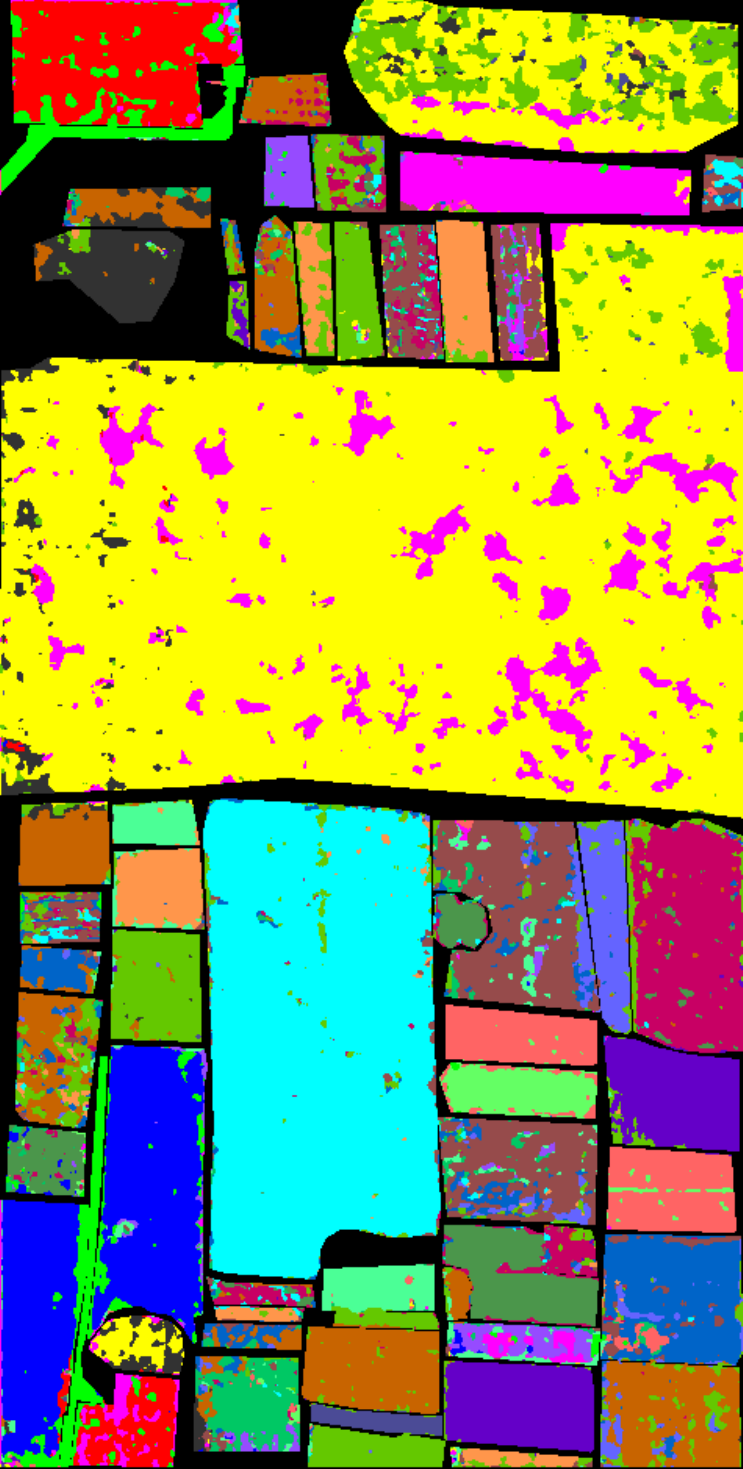}
\caption{}
\label{fig:SSFTT_HH}
\end{subfigure}
\\
\begin{subfigure}{.12\textwidth}
\centering
\includegraphics[width=1\linewidth]{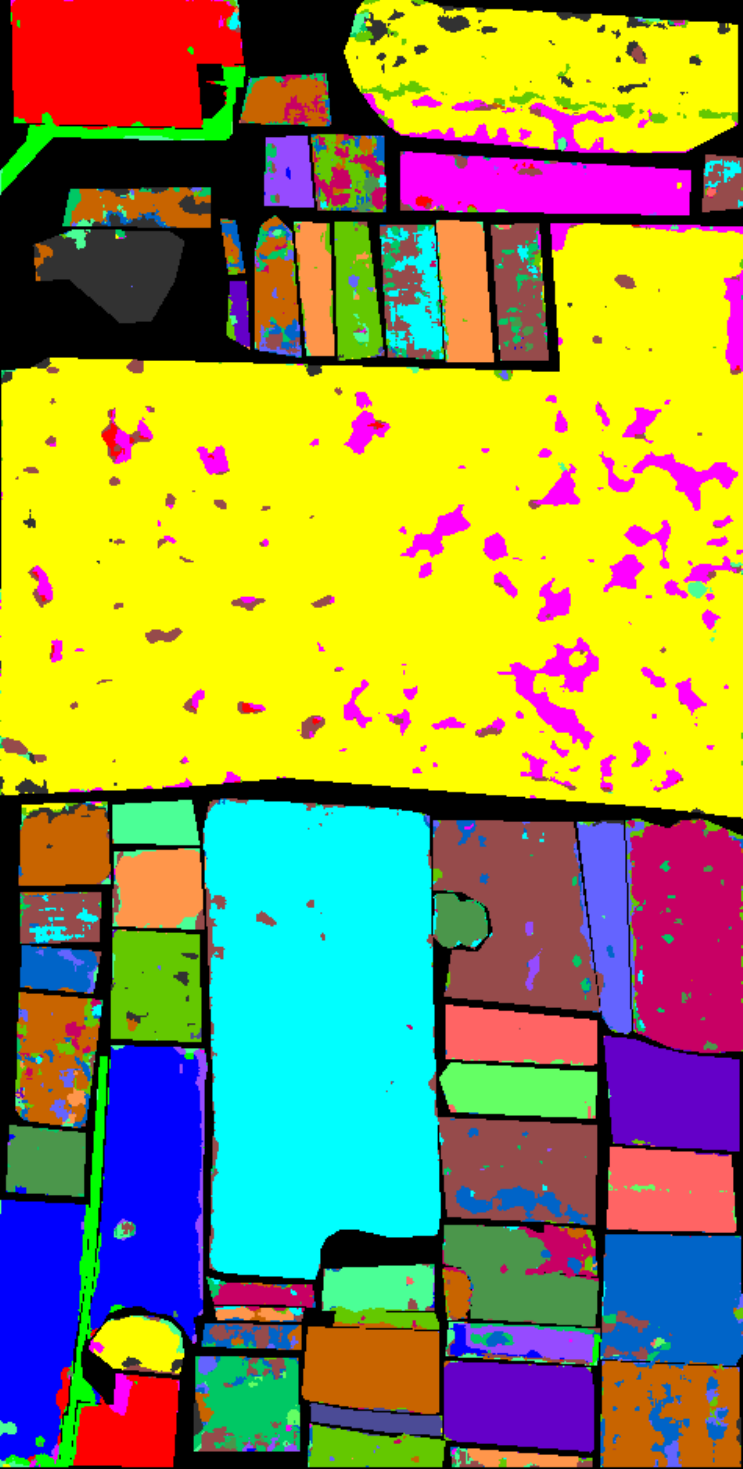}
\caption{}
\label{fig:MF_HH}
\end{subfigure}
\begin{subfigure}{.12\textwidth}
\centering
\includegraphics[width=1\linewidth]{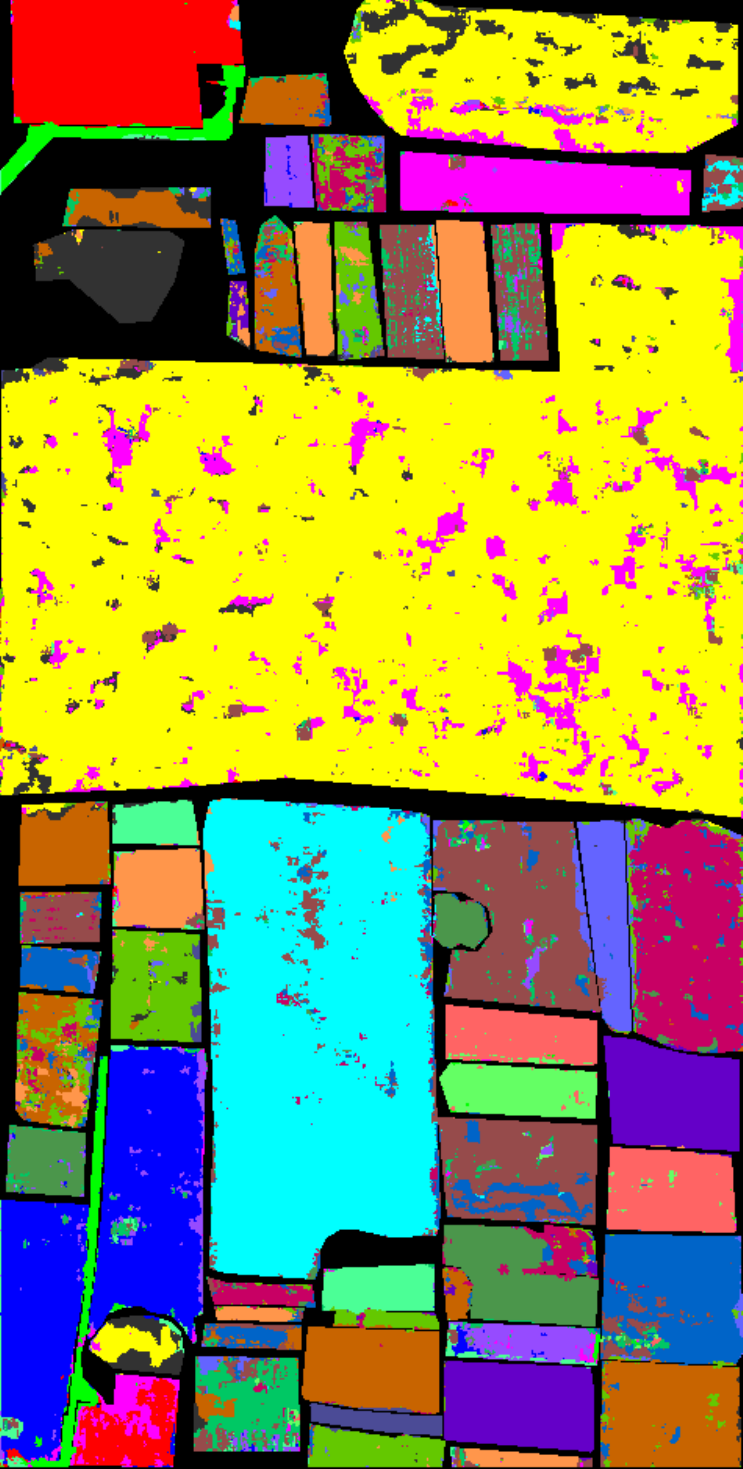}
\caption{}
\label{fig:GSC_HH}
\end{subfigure}
\begin{subfigure}{.12\textwidth}
\centering
\includegraphics[width=1\linewidth]{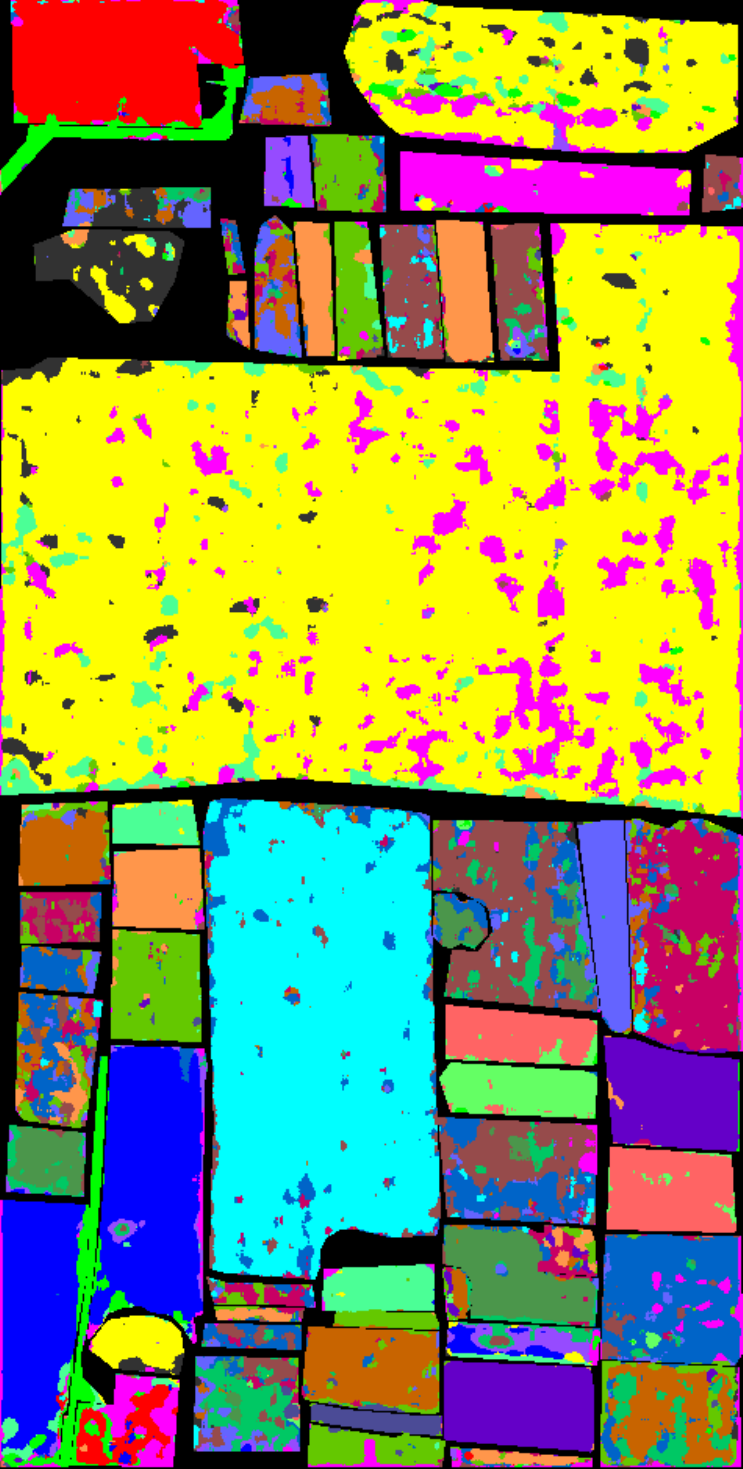}
\caption{}
\label{fig:TDSS_HH}
\end{subfigure}
\begin{subfigure}{.12\textwidth}
\centering
\includegraphics[width=1\linewidth]{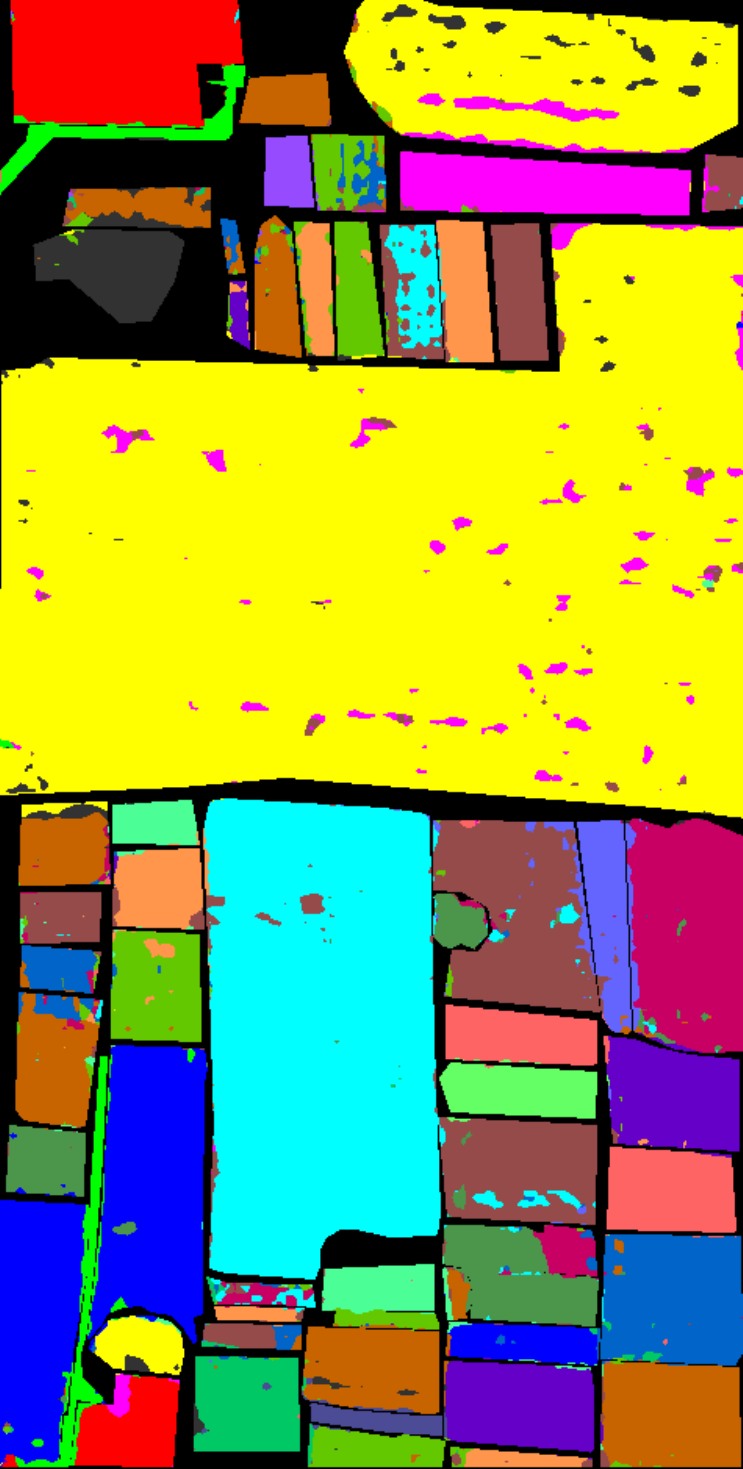}
\caption{}
\label{fig:MambaHSI_HH}
\end{subfigure}
\begin{subfigure}{.12\textwidth}
\centering
\includegraphics[width=1\linewidth]{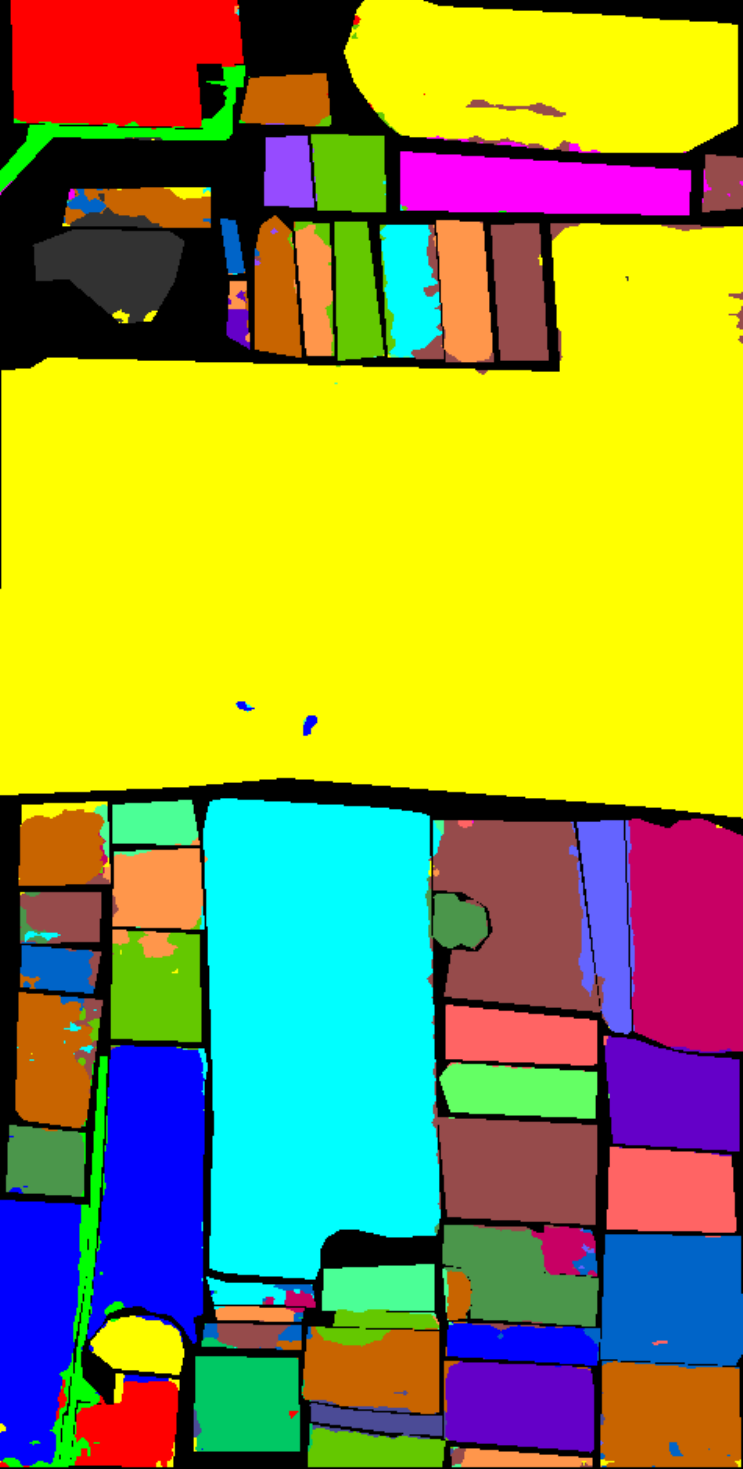}
\caption{}
\label{fig:proposed_HH}
\end{subfigure}
\caption{Visualization of the classification results for WHU-Hi-HongHu dataset. (a) Ground-truth map. (b) SVM. (c) 3D-CNN. (d) FullyContNet. (e) SSFTT. (f) MorpyFormer. (g) GSC-ViT. (h) 3DSS-Mamba. (i) MambaHSI. (j) proposed HS-Mamba.}
\label{fig:visualization_HH}
\end{figure*}

As shown in Table \ref{tab:contrastHH}, our proposed HS-Mamba demonstrates robust performance on the large-resolution HSI dataset HH. Specifically, HS-Mamba surpasses the second-best method MambaHSI by margins of 2.30\%, 1.57\%, and 2.84\% in OA, AA, and Kappa metrics, respectively. As shown in Fig.  \ref{fig:visualization_HH}, large-area crops such as Cotton and Rape exhibit smoother and more accurate classification results compared to other methods, demonstrating the potential of our approach for large-resolution HSI analysis.

\subsection{Ablation Study}
\subsubsection{Effect of dual domains design}
As shown in Table \ref{tab:abl_dual_domains}, using either the spatial or spectral branch alone achieves strong performance, with the spatial branch performing slightly better, demonstrating superior precision in spatial feature perception and enhanced feature capture capability. Combining both domains through gated fusion further boosts accuracy, as the strategy strategically integrates complementary spatial-spectral representations to mitigate individual domain limitations, highlighting their synergistic roles and necessity for optimal results.
\begin{table}[b]
\renewcommand\arraystretch{1}
\centering
\caption{Ablation study of the proposed dual-domain framework, using Overall Accuracy (OA\%) as the metric. Spa and Spe denote the spatial-domain branch and spectral-domain branch, respectively.}
\label{tab:abl_dual_domains}
\begin{tabular}{cc|cccc}
\hline
\hline
Spe & Spa & IP & PU & HC & HH\\
\hline
\checkmark & & 93.58±1.14 & 94.59±1.49 & 95.35±0.84 & 94.51±0.68\\
& \checkmark & 94.41±1.12 & 95.87±1.14 & 95.83±0.82 & 94.67±0.96\\
\checkmark & \checkmark & \textbf{94.60±1.28} & \textbf{96.35±1.56} & \textbf{95.93±0.74} & \textbf{95.25±0.66}\\
\hline
\hline
\end{tabular}
\end{table}

\subsubsection{Effect of the position encoding}
As shown in Table \ref{tab:abl_pos_encoding}, the quantitative results demonstrate the benefits of incorporating positional encoding, with accuracy improvements of 2\% to 7\% observed consistently across all datasets. By recurrently integrating positional encoding into each DCSS-Encoder layer, the repeated emphasis on positional information enables enhanced awareness of non-overlapping patches' spatial relationships, ultimately achieving promising performance improvements.
\begin{table}[htb]
\renewcommand\arraystretch{1}
\centering
\caption{Ablation study of the proposed method in fusion strategy, OA(\%) is used as the indicator. }
\label{tab:abl_pos_encoding}
\begin{tabular}{c|cccc}
\hline
\hline
Pos. Encoding & IP & PU & HC & HH\\
\hline
 & 92.28±1.66 & 89.57±1.11 & 92.53±2.36 & 92.86±0.74\\
\checkmark & \textbf{94.60±1.28} & \textbf{96.35±1.56} & \textbf{95.73±0.74} & \textbf{95.25±0.66}\\
\hline
\hline
\end{tabular}
\end{table}

\subsubsection{Effect of the full-field interaction strategy}
As shown in Table \ref{tab:abl_full_field}, the quantitative experiments demonstrate that integrating the LGI Attention branch effectively incorporates comprehensive global context into localized features, achieving mean OA improvements of 0.1\% to 1\% across four datasets while reducing standard deviations by approximately 0.3\%. These results confirm the module's dual capability for enhanced accuracy and stabilized predictions, ultimately validating the necessity of attention mechanisms in our architecture.

\begin{table*}[t]
\renewcommand\arraystretch{1}
\centering
\caption{Ablation study of the full-field interaction strategy, using Overall Accuracy and FLOPs as the metric.}
\label{tab:abl_full_field}
\begin{tabular}{cc|cc|cc|cc|cc}
\hline
\hline
\multirow{2}{*}{DCSS-Encoder}&
\multirow{2}{*}{LGI-Atten}&
\multicolumn{2}{c|}{IP}&
\multicolumn{2}{c|}{PU}&
\multicolumn{2}{c|}{HC}&
\multicolumn{2}{c}{HH}\\
\cline{3-10}
& & OA(\%) & FLOPs(G) & OA(\%) & FLOPs(G) & OA(\%) & FLOPs(G) & OA(\%) & FLOPs(G)\\
\hline
\checkmark & & 94.60±1.28 & 3.23 & 96.35±1.56 & 25.45 & 95.73±0.74 & 61.26 & 95.25±0.66 & 79.39\\
\checkmark & \checkmark & \textbf{94.67±0.91} & 3.24 & \textbf{96.43±1.35} & 25.53 & \textbf{96.34±0.79} & 61.55 & \textbf{96.25±0.37} & 79.56\\
\hline
\hline
\end{tabular}
\end{table*}

\subsubsection{Effect of the fusion strategy}
As shown in Table \ref{tab:abl_fusion}, the results show the effect of various fusion strategies. The proposed Gated Fusion performs best on IP, HC, and HH datasets, achieving higher accuracy and lower standard deviation. While sum and adaptive sum slightly outperform it on PU, gated fusion remains competitive and still surpasses the concat fusion in this scenario. Overall, gated fusion proves more robust across datasets, demonstrating broader applicability than the three fusion way in most scenarios.

\subsection{Analysis of Hyper-parameters}
\subsubsection{Effect of patch size}
\begin{figure}
\centering
\begin{subfigure}{.24\textwidth}
\centering
\includegraphics[width=1\linewidth]{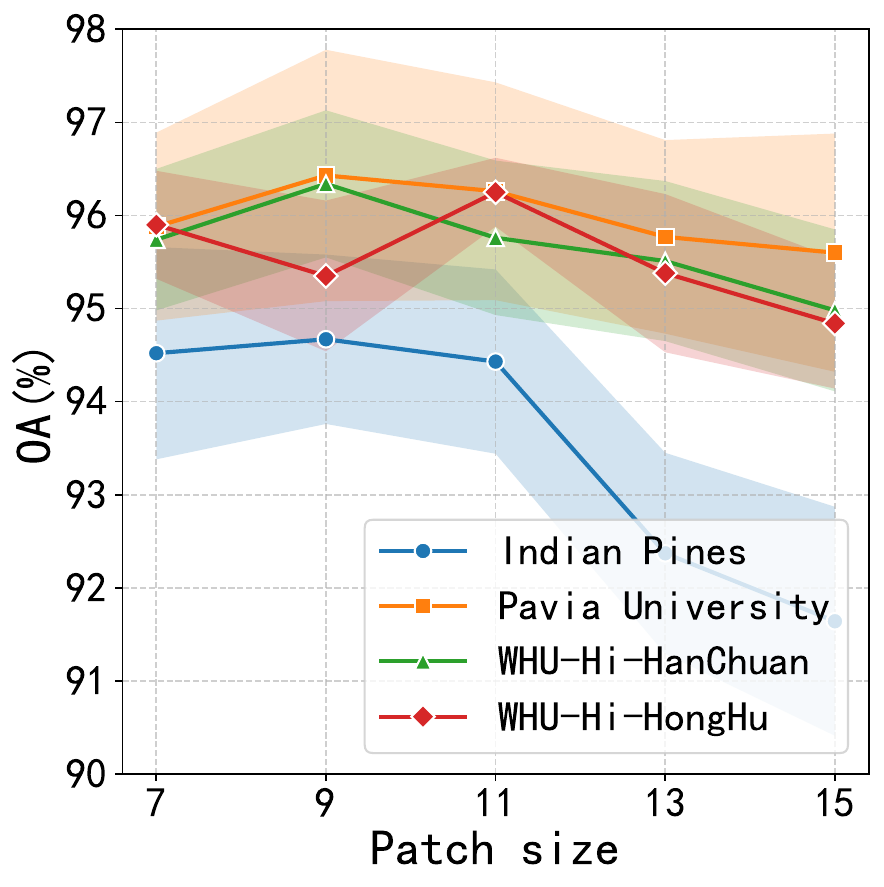}
\caption{}
\label{fig:experiment_patch}
\end{subfigure}
\begin{subfigure}{.24\textwidth}
\centering
\includegraphics[width=1.\linewidth]{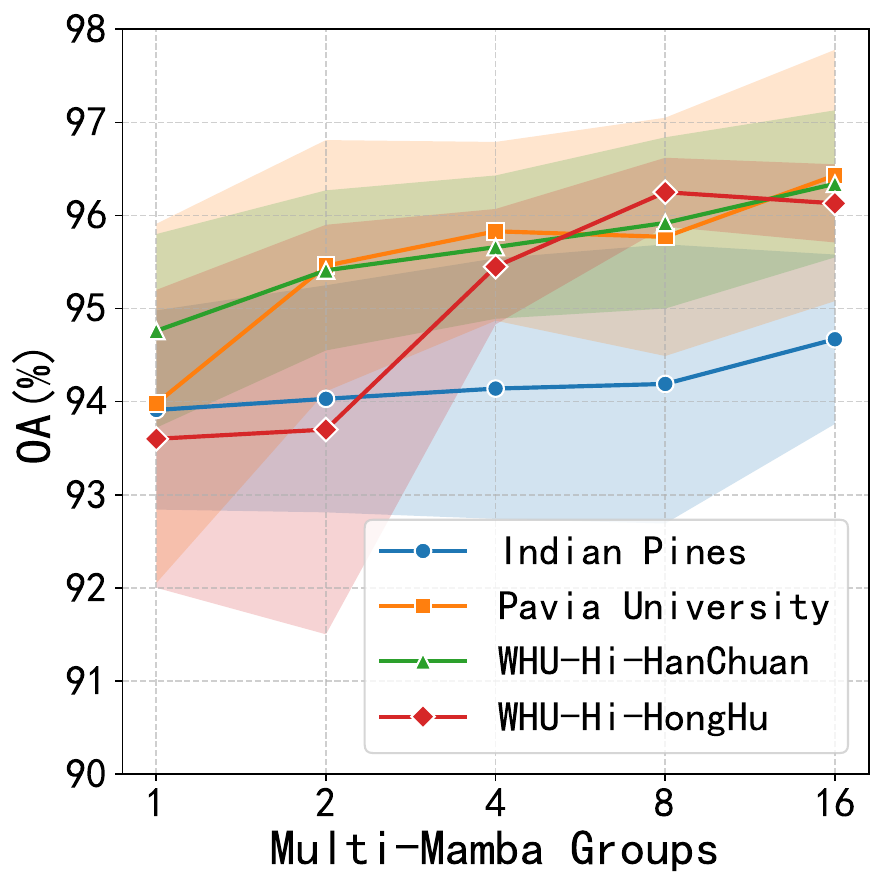}
\caption{}
\label{fig:experiment_group}
\end{subfigure}
\caption{Hyper-parameter analysis for four datasets. (a) Effect of patch size. (b) Effect of mamba groups. The figure reports the mean and standard deviation of the results from ten runs.}
\label{fig:hyper_param_experiment}
\end{figure}

As shown in Fig. \ref{fig:experiment_patch}, we analyzed the patch size across four datasets. Classification performance improved for all datasets when patch size increased from 7 to 9. However, further increases beyond 9 caused significant performance degradation for IP, PU, and HC, while HH peaked at 11 before declining. This suggests that smaller patch sizes better capture fine-grained local features when processing non-overlapping patches.

\begin{table}[H]
\renewcommand\arraystretch{1}
\centering
\caption{Ablation study of the proposed method in fusion strategy, OA(\%) is used as the indicator. }
\label{tab:abl_fusion}
\begin{tabular}{c|cccc}
\hline
\hline
Fusion Way & IP & PU & HC & HH\\
\hline
Sum & 93.91±1.61 & \textbf{96.72±0.89} & 95.94±0.93 & 95.39±0.72\\
Adaptive Sum & 94.08±1.41 & 96.71±0.89 & 95.95±0.93 & 95.45±0.61\\
Concat Fusion & 94.60±1.24 & 95.37±2.11 & 95.82±0.91 & 95.12±0.61\\
Gated Fusion & \textbf{94.67±0.91} & 96.43±1.35 & \textbf{96.34±0.79} & \textbf{96.25±0.37}\\
\hline
\hline
\end{tabular}
\end{table}

\subsubsection{Effect of multi-mamba groups}
As shown in Fig. \ref{fig:experiment_group}, we conducted hyperparameter analysis of multi-mamba groups across four datasets. The IP, PU, and HC datasets achieved optimal performance with groups equal to 16, while the HH dataset peaked at groups equal to 8. This indicates varying dataset sensitivities to spatial-spectral group configurations, with HH's large resolution and broad land cover distribution requiring fewer groups for sufficient modeling.

\subsection{Analysis of Computational Complexity}
We evaluate computational efficiency across four datasets by comparing training time, inference time, parameters, and FLOPs between our method and SOTA approaches. As shown in Table \ref{tab:Computational_Complexity}, pixel-patch methods achieve 36\% faster training times, while whole-image approaches and our full-field interaction strategy demonstrate significantly faster inference speeds. This stems from redundant computations in pixel-patch methods compensating for limited global context modeling, where the requirement to process every pixel individually during inference drastically increases operational complexity, leading to higher FLOPs and slower inference compared to whole-image strategies\cite{MambaHSI}.

Our method achieves 52\% faster inference and 26\% lower FLOPs than MambaHSI on medium-high resolution datasets, with performance advantages amplifying at larger resolutions. Quantitative analysis of metrics further demonstrates that our full-field interaction strategy achieves superior classification accuracy with lower computational demands compared to other approaches across all four datasets, maintaining reduced parameters and FLOPs while delivering SOTA performance. This evidence confirms the computational superiority of our proposed strategy.

\begin{table*}[htb!]
\centering
\caption{Quantitative computational efficiency metrics (Training time, Inference time, Parameters and FLOPs) for four datasets.}
\label{tab:Computational_Complexity}
\begin{tabular}{c|c|c|cc|ccc|ccc}
\hline  
\hline
\multirow{3}{*}{Dataset}& \multirow{3}{*}{Times}& \multicolumn{1}{c|}{ML-based} & \multicolumn{2}{c|}{CNN-based} & \multicolumn{3}{c|}{Transformer-based} & \multicolumn{3}{c}{Mamba-based}\\\cline{2-11}
& & SVM & 3D-CNN & FullyContNet & SSFTT & MorphFormer & GSC-ViT & 3DSS-Mamba & MambaHSI & HS-Mamba\\
& & {\tiny TGRS2004} & {\tiny TGRS2020} & {\tiny TGRS2022} & {\tiny TGRS2022} & {\tiny TGRS2023} & {\tiny TGRS2024} & {\tiny TGRS2024} & {\tiny TGRS2024} & {\tiny ours}\\
\hline
\multirow{4}{*}{IP} & $T_{tr}(s)$ & - & 32.24 & 17.32 & 22.23 & 64.95 & 34.85 & 159.98 & 59.26 & 88.44\\
 & $T_{in}(s)$ & - & 1.45 & 0.01 & 1.04 & 5.75 & 2.13 & 6.61 & 0.03 & 0.11\\
 & Params(K) & - & 263.60 & 1370.06 & 936.58 & 277.55 & 638.48 & 24.37 & 425.56 & 188.81\\
 & FLOPs(G) & - & 1910.00 & 17.95 & 690.46 & 831.12 & 897.77 & 1840.00 & 4.67 & 3.24\\
\hline
\multirow{4}{*}{PU} & $T_{tr}(s)$ & - & 12.11 & 81.27 & 11.95 & 33.60 & 19.23 & 55.12 & 455.83 & 309.90\\
 & $T_{in}(s)$ & - & 6.67 & 0.04 & 6.22 & 44.13 & 13.18 & 33.33 & 0.35 & 0.50\\
 & Params(K) & - & 101.19 & 1310.00 & 489.15 & 226.86 & 152.71 & 24.14 & 412.24 & 177.26\\
 & FLOPs(G) & - & 9710.00 & 155.62 & 3510.00 & 5150.00 & 2030.00 & 10960.00 & 41.04 & 25.53\\
\hline
\multirow{4}{*}{HC} & $T_{tr}(s)$ & - & 52.90 & 46.58 & 33.00 & 81.61 & 46.15 & 100.17 & 829.51 & 495.30\\
 & $T_{in}(s)$ & - & 32.65 & 0.08 & 23.93 & 119.44 & 34.84 & 60.98 & 0.65 & 0.85\\
 & Params(K) & - & 344.24 & 1410.00 & 1280.00 & 336.75 & 643.34 & 24.37 & 435.03 & 199.77\\
 & FLOPs(G) & - & 46000.00 & 348.58 & 16590.00 & 19690.00 & 15970.00 & 19490.00 & 89.08 & 61.65\\
\hline
\multirow{4}{*}{HH} & $T_{tr}(s)$ & - & 105.86 & 146.51 & 43.48 & 114.15 & 62.58 & 134.73 & 657.07 & 338.03\\
 & $T_{in}(s)$ & - & 39.14 & 0.09 & 27.75 & 145.62 & 36.81 & 70.87 & 0.77 & 0.65\\
 & Params(K) & - & 449.52 & 1410.00 & 1260.00 & 333.94 & 642.71 & 24.57 & 435.29 & 236.64\\
 & FLOPs(G) & - & 54960.00 & 420.75 & 19790.00 & 23510.00 & 19280.00 & 23600.00 & 107.50 & 79.56\\
\hline
\hline
\end{tabular}
\end{table*}

\section{Conclusion}
In this paper, we propose HS-Mamba, a multi-group Mamba framework with full-field interaction architecture for HSI classification. This method effectively integrates the advantages of pixel-patch strategies and whole-image approaches, enabling comprehensive global-local feature interaction to efficiently capture spectral-spatial information. Extensive evaluations on four public HSI datasets demonstrate that HS-Mamba achieves superior classification performance while maintaining high computational efficiency, outperforming state-of-the-art (SOTA) methods.

\FloatBarrier
\begin{refcontext}[sorting = none]
\printbibliography
\end{refcontext}

\end{document}